\documentclass[final,3p,11pt,times]{elsarticle}

\usepackage{booktabs}
\usepackage{multirow}
\usepackage{makecell}
\usepackage{subfigure}
\usepackage{bm}
\usepackage{amsmath}
\usepackage{xcolor}
\usepackage[normalem]{ulem}
\useunder{\uline}{\ul}{}
\usepackage[colorlinks,
            linkcolor=blue,
            anchorcolor=blue,
            citecolor=blue]{hyperref}

\usepackage{amssymb}
\usepackage{amsthm}
\newtheorem{definition}{Definition}
\journal{Artificial Intelligence}

\begin{document}

\begin{frontmatter}



\title{Temporal Inductive Path Neural Network for Temporal Knowledge Graph Reasoning}


\author[1,2]{Hao Dong\corref{ra}}
\cortext[ra]{This work was done when the first-author was the research assistant in IOTSC at University of Macau.}
\ead{donghcn@gmail.com}

\author[3]{Pengyang Wang\corref{cor1}}
\cortext[cor1]{Corresponding authors.}
\ead{pywang@um.edu.mo}

\author[1,2]{Meng Xiao}
\ead{shaow@cnic.cn}

\author[1,2]{Zhiyuan Ning}
\ead{ningzhiyuan@cnic.cn}

\author[1,2]{Pengfei Wang\corref{cor1}}
\ead{pfwang@cnic.cn}

\author[1,2]{Yuanchun Zhou}
\ead{zyc@cnic.cn}

\affiliation[1]{%
  organization={Computer Network Information Center, Chinese Academy of Sciences},
  addressline={2 Dongsheng South Rd},
  city={Haidian District},
  state={Beijing},
  country={China}}
\affiliation[2]{%
  organization={University of Chinese Academy of Sciences},
  addressline={1 Yanqihu East Rd},
  city={Huairou District},
            state={Beijing},
  country={China}}
\affiliation[3]{%
  organization={Department of Computer and Information Science, The State Key Laboratory of Internet of Things for Smart City, University of Macau},
  addressline={Avenida da Universidade},
  city={Taipa},
  state={Macau},
            country={China}}

\begin{abstract}
    Temporal Knowledge Graph (TKG) is an extension of traditional Knowledge Graph (KG) that incorporates the dimension of time. Reasoning on TKGs is a crucial task that aims to predict future facts based on historical occurrences. 
    The key challenge lies in uncovering structural dependencies within historical subgraphs and temporal patterns. 
    Most existing approaches model TKGs relying on entity modeling, as nodes in the graph play a crucial role in knowledge representation.
    However, the real-world scenario often involves an extensive number of entities, with new entities emerging over time. This makes it challenging for entity-dependent methods to cope with extensive volumes of entities, and effectively handling newly emerging entities also becomes a significant challenge. 
    Therefore, we propose \textbf{T}emporal \textbf{I}nductive \textbf{P}ath \textbf{N}eural \textbf{N}etwork (TiPNN),
    which models historical information in an entity-independent perspective. 
    Specifically, TiPNN adopts a unified graph, namely history temporal graph, to comprehensively capture and encapsulate information from history. 
    Subsequently, we utilize the defined query-aware temporal paths on a history temporal graph to model historical path information related to queries for reasoning.
    % TiPNN models historical information without relying on entity representations, allowing for natural handling of the inductive setting in TKG scenarios. 
    % Extensive experiments demonstrate that the proposed model achieves substantial performance improvements and natural handling of the inductive setting while also enabling the provision of reasoning evidence through history temporal graphs.
    Extensive experiments illustrate that the proposed model not only attains significant performance enhancements but also handles inductive settings, while additionally facilitating the provision of reasoning evidence through history temporal graphs.
\end{abstract}


\end{frontmatter}


% \section{}
% \label{}
\section{Introduction}

Knowledge Graphs (KGs) are powerful representations of structured information that capture a vast array of real-world facts. 
They consist of nodes, which represent entities such as people, places, and events, and edges that define the relationships between these entities. 
These relationships are typically represented as triples in the form of $(s, r, o)$, comprising a subject, a relation, and an object, such as \textit{(Tesla, Founded by, Elon Musk)}. 
Temporal Knowledge Graphs (TKGs) is an extension of a traditional KGs that incorporates the dimension of time. In a TKG, these facts are annotated with timestamps or time intervals to indicate when they are or were valid. 
Different from KGs, each fact in a TKG is represented as a quadruple, adding a timestamp feature, which can be denoted as $(s, r, o, t)$, including a subject, a relation, an object, and a timestamp, 
such as \textit{(Albert Einstein, Born in, Germany, 03/14/1879)}. 
The timestamp indicates the specific date or time when this event occurred, providing temporal information to the graph. 
Such unique representation empowers TKGs to capture the dynamics of multi-relational graphs over time, exhibiting temporal variations. 
As a result, TKGs have found widespread applications in diverse domains, like social network analysis~\cite{a272db48509f498facf752cb886353a9,doi:10.1177/1525822X16643709}, and event prediction~\cite{jin2019recurrent,10.1145/3443687,DBLP:conf/kdd/WangLJLF20} (e.g. disease outbreaks, natural disasters, and political shifts).
These applications leverage the temporal information embedded in TKGs to make informed decisions and predictions based on historical knowledge. 

Among various tasks, TKG reasoning focuses on inferring missing facts based on known ones and consists of two primary settings: interpolation and extrapolation~\cite{jin2020Renet}. 
Given the timestamp scope $[0,T]$ in a TKG, where $T$ represents the maximum timestamp,
in the interpolation setting, it focuses on the time range from $0$ to $T$, i.e., completing missing facts within the range of past timestamps~\cite{2020Diachronic}. On the other hand, in the extrapolation setting, the reasoning timestamps extend beyond $T$, focusing on predicting the facts in the future~\cite{li2021temporal,dong2023adaptive,2020Learning}.
This study particularly emphasizes the extrapolation setting.

Existing literature on TKG reasoning mainly focuses on exploring temporal features and structural information to capture complicated variations and logical connections between entities and relations. 
% Historical sequential data encapsulates the developmental patterns of facts within a given context, the logical connections between entities and relations, and the temporal variations. Unearthing comprehensive historical features from sequential data is crucial for predicting missing facts about the future~\cite{sanchez2020learning}.
% Currently, there are numerous studies that explore temporal features and structural information in TKGs.
% By leveraging various sequence modeling techniques, such as RNNs, LSTMs, and gate-based methods, it aims to extract hidden temporal patterns from TKGs, learn dynamic characteristics of subgraphs at different timestamps, and capture the evolutionary trends of entities and relations. 
% When considering structural features in TKGs, the relationships between entities and the interactions between entities and relations are crucial factors to be taken into account. Graph Neural Network (GNN) methods have been widely applied to model structural features in TKGs. These methods effectively capture the dependencies and patterns in the graph structure, enabling the representation learning of entities and relations in a meaningful way. 
For example, RE-NET~\cite{jin2020Renet} and RE-GCN~\cite{li2021temporal} are representative works that leverage RNN-based and GNN-based approaches to learn temporal and structural features of entities in TKGs. xERTE~\cite{han2020explainable} utilizes the entity connection patterns from a constructed inference graph. These methods rely on entity modeling as nodes in the graph play a crucial role in knowledge representation. 
However, the real-world scenario often involves an extensive number of entities, with new entities emerging over time. This makes it challenging for entity-dependent methods to cope with extensive volumes of entities, and effectively handling newly emerging entities also becomes a significant challenge.
% Considering the growing number of new entities emerging over time and the vast number of entities in the real world, 

To address this, we previously introduced an innovative \textit{entity-independent} modeling approach for TKGs with an evolutionary perspective in our previous work, referred to as DaeMon~\cite{dong2023adaptive}.
The entity-independent modeling approach brings several benefits: 
(i) modeling on graph representation is independent of entities, maintaining stable performance even for datasets with numerous entities. Besides, the memory occupation is independent of the number of entity categories.
(ii) the entity-independent nature makes it insensitive to entity-specific features. Hence, when new entities emerge in future timestamps, the model can still handle these unseen nodes, showcasing robust generalization and reasoning capabilities.
In essence, DaeMon focuses on learning query-aware latent path representations, incorporating a passing strategy between adjacent timestamps to control the path information flow toward later ones without explicitly modeling individual entities.
Specifically, DaeMon introduces the concept of temporal paths to represent the logical paths between a query's subject entity and candidate object entities within the sequence of historical subgraphs. The model employs a path aggregation unit to capture local path information within each subgraph
and utilizes path memory units to store aggregated representations of temporal paths along the timeline.  Finally, DaeMon adopts a memory passing strategy to facilitate cross-temporal information propagation.

\begin{figure}[t]
    \centering
    \includegraphics[width=0.85\textwidth]{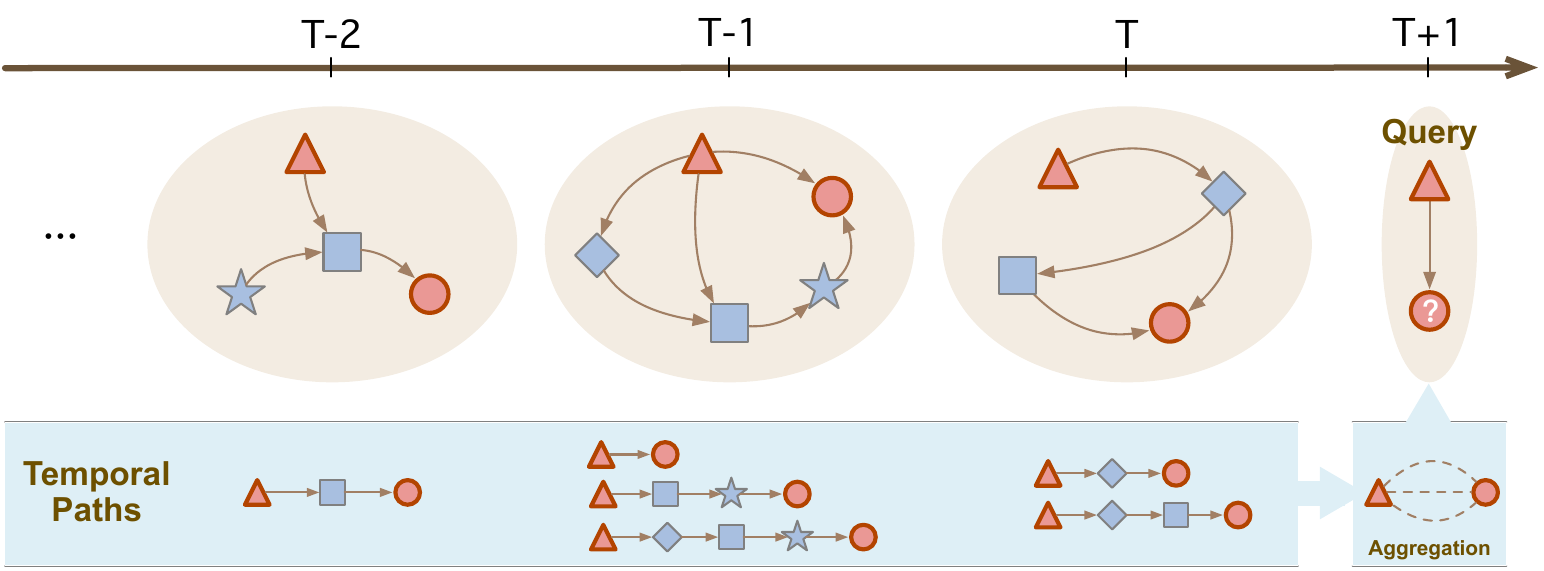}
    \caption{An example of inference based on temporal paths modeling. The red triangular nodes and red circular nodes represent the subject entity and candidate object entity of the query at timestamp $T+1$, respectively.}
    \label{fig:intro}

\end{figure}

Here, we provide a generalized overview of inference based on temporal paths modeling, as illustrated in Figure~\ref{fig:intro}.
For a given inference query, it starts from this query and collects the path between the subject entity and object entity in the historical context. Then, it utilizes a query-aware temporal method to represent the aggregation of temporal paths with temporal information for reasoning. 
For example, as shown in Figure~\ref{fig:intro}, it iteratively learns the path information between the red triangular and red circular nodes within the historical context. The aggregation of path information corresponding to the query is then utilized to infer the possibility of potential future interactions. Note that, during the learning of path information, the method does not focus on which intermediary nodes (i.e., the blue nodes in Figure~\ref{fig:intro}) are present in the path, but rather employs the relation representations present in the path to express it, showcasing an entity-independent characteristic.

While DaeMon benefits from this entity-independent modeling approach, it still exhibits certain limitations.
First, during the learning of historical information, DaeMon independently performs graph learning on local subgraphs and then connects them through memory passing strategy to obtain the final query-aware path features for future fact reasoning. This process requires graph learning on each subgraph separately, leading to difficulties in modeling complex temporal characteristics and a significant increase in time complexity as the historical length grows.
Second, the memory passing strategy is adopted to model temporal features in historical sequences, controlling the information flow from one timestamp to the next to achieve temporal evolution modeling. However, this temporal modeling approach is unidirectional, which not only gives rise to the long-term dependency problem but also neglects relative temporal features between timestamps.
Third, DaeMon defines temporal paths as virtually connected paths across the sequence of subgraphs. In practice, it independently learns several edges on each timestamp's subgraph and then obtains the path feature by fusing the edge information through memory passing strategy. The independent learning approach without real physical paths creates challenges in providing relevant reasoning evidence during the inference stage, which makes it difficult to present interpretable explanations to users.

To overcome the above challenges, here we propose a Temporal Inductive Path Neural Network (TiPNN). Specifically, we introduce the concept of \textbf{\textit{history temporal graphs}} as a replacement for the original historical subgraph sequences to mitigate the complexity of learning historical information independently on multiple subgraphs. It allows us to model historical information more efficiently by combining it into a single unified graph.
% , thereby enabling the discovery of complex temporal characteristics and reducing the computational burden during model inference.
Furthermore, we redefine the notion of \textit{\textbf{temporal paths}} as a logical semantics path between the query entity and candidate entities in history temporal graph and design \textbf{\textit{query-aware temporal path processing}} to model the path feature on the single unified graph in the entity-independent manner for the final reasoning. 
It enables us to comprehensively learn both the temporal and structural features present in the history temporal graph. 
The adoption of the entity-independent approach allows the model to effectively handle the inductive setting on the history temporal graph. By jointly modeling the information within and between different timestamps in a single unified graph, it can capture a more intuitive and holistic view of the temporal reasoning logic, thus facilitating user understanding of the model's reasoning process and can even provide interpretable explanations for the reasoning results. Our main contributions are as follows:
\begin{itemize}
\item\noindent To holistically capture historical connectivity patterns, we introduce a unified graph, namely the history temporal graph, to retain complete features from the historical context. The history temporal graph integrates the relationship of entities, timestamps of historical facts, and temporal characteristics among historical facts.
\item\noindent We define the concept of temporal paths and propose the Temporal Inductive Path Neural Network (TiPNN) for the task of extrapolated reasoning over TKGs. Our approach models the temporal path features corresponding to reasoning queries by considering temporal information within and between different timestamps, along with the compositional logical rules of historical subgraph sequences.
\item\noindent Extensive experiments demonstrate that our model outperforms the existing state-of-the-art results. Additionally, we construct datasets and conduct validation experiments for inductive reasoning on TKGs. Furthermore, we present reasoning evidence demonstrations and analytical experiments based on history temporal graph inference.
\end{itemize}

\section{Preliminaries}

In this section, we provide the essential background knowledge and formally define the context. We also introduce the basics of temporal knowledge graph reasoning task. The details of essential mathematical symbols and the corresponding descriptions of TiPNN are shown in Table \ref{tab:notation}. It is important to clarify beforehand that we will indicate vector representations in the upcoming context using \textbf{bold} items.

\begin{table}[!htbp]
    \centering
    % \footnotesize
    % \setlength\tabcolsep{12pt}
    \caption{Notations and Descriptions.}
    \begin{tabular}{ll}
    \toprule
    Notations & Descriptions  \\ 
    \midrule

    $\mathcal{G}$   &   a temporal knowledge graph  \\ 
    $G_t$   &   a subgraph corresponding to the snapshot at timestamp $t$  \\ 
    $\mathcal{V}, \mathcal{R}, \mathcal{T}$   &  the finite set of entities, relation types and timestamps in the temporal knowledge graph   \\
    $\mathcal{E}_t$ &   the set of fact edges at timestamp $t$  \\
    $(s,r,o,t)$  &  a quadruple (fact edge) at timestamp $t$  \\
    $(s,r,?,t+1)$   &   a query with the missing object that is to be predicted at timestamp $t+1$ \\
    % \midrule
    
    \midrule
    $\hat{G}$ &   a history temporal graph  \\
    $r_\tau$    &   a temporal relation in history temporal graph   \\
    $(s, r_\tau, o)$    &   a temporal edge with time attribute $\tau$ attached in history temporal graph   \\
    $m$ &   the length of the history used for reasoning    \\
    $\omega$    &   the number of temporal path aggregation layers \\
    $\textbf{H}$, $\textbf{h}$    &   the representation of temporal paths (and path)  \\
    $\textbf{R}$    &   a learnable representation of relation types in $\mathcal{R}$  \\
    $\bm{\Psi}_r$    &   the query relation $r$-aware basic static embedding of temporal edge   \\
    $\bm{\Upsilon}$ &   the temporal embedding of temporal edge  \\
    
    $\textbf{w}_r$  &   the query-aware temporal representation of a temporal edge  \\
    $d$ &   the dimension of embedding  \\
    
    \bottomrule
    \end{tabular}
    
    % \vspace{-4mm}
    \label{tab:notation}
\end{table}

\subsection{Background of Temporal Knowledge Graph}

A temporal knowledge graph $\mathcal{G}$ is essentially a multi-relational, directed graph that incorporates timestamped edges connecting various entities. It can be formalized as a succession of static subgraphs arranged in chronological order, i.e., $\mathcal{G}=\{G_1, G_2,..., G_t,...\}$.
A subgraph in $\mathcal{G}$ can be denoted as $G_t=(\mathcal{V}, \mathcal{R}, \mathcal{E}_t)$ corresponding to the snapshot at timestamp $t$, where $\mathcal{V}$ is the set of entities, $\mathcal{R}$ is the set of relation types, and $\mathcal{E}_t$ is the set of fact edges at timestamp $t$. Each element in $\mathcal{E}_t$ is a timestamped quadruple $(subject, relation, object, timestamp)$, which can be denoted as $(s,r,o,t)$ or $(s_t,r_t,o_t)$, describing a relation type $r \in \mathcal{R}$ occurs between subject entity $s \in \mathcal{V}$ and object entity $o \in \mathcal{E}$ at timestamp $t \in \mathcal{T}$, where $\mathcal{T}$ denote the finite set of timestamps. 

\subsection{Formulation of Reasoning Task}

Temporal knowledge graph reasoning task involves predicting the missing entity by answering a query like $(s,r,?,t_q)$ using historical known facts $\{(s,r,o,t_i)|t_i<t_q\}$. Note that the facts in the query time period $t_q$ are unknown in this task setting.
For the sake of generalization, we assume that the fact prediction at a future time depends on a sequence of historical subgraphs from the closest $m$ timestamps. That is, in predicting the missing fact at timestamp $t+1$, we consider the historical subgraph sequence $\{G_{t-m+1},..., G_t\}$ for the inference. 

In addition, given an object entity query $(s,r,?,t_q)$ at a future timestamp $t_q$, we consider all entities in the entity set $\mathcal{V}$ as candidates for the object entity reasoning. The final prediction is obtained after scoring and ranking all the candidates by a scoring function. When predicting the subject entity, i.e., $(?,r,o,t_q)$, we can transform the problem into object entity prediction form $(o,r^{-1},?,t_q)$. Therefore, we also insert the corresponding reverse edge $(o,r^{-1},s,t)$ when processing the history graph. Without loss of generality, the later section will be introduced in terms of object entity predictions.

\section{Methodology}

In this section, we introduce the proposed model, \textbf{T}emporal \textbf{I}nductive \textbf{P}ath \textbf{N}eural \textbf{N}etwork (TiPNN). We start with a model overview and then discuss each part of the model as well as its training and reasoning process in detail. 

\subsection{Model Overview}

The main idea of TiPNN is to model the compositional logical rules on multi-relational temporal knowledge graphs. By integrating historical subgraph information and capturing the correlated patterns of historical facts, we can achieve predictions for missing facts at future timestamps. 
Given a query $(s,r,?,t+1)$ at future timestamp $t+1$, our focus is on utilizing historical knowledge derived from $\{G_{t-m+1},..., G_t\}$ to model the connected semantic information between the query subject $s$ and all candidate objects $o \in \mathcal{V}$ in history for the query responding. 

Each subgraph in a temporal knowledge graph with a different timestamp describes factual information that occurred at different moments. These subgraphs are structurally independent of each other, as there are no connections between the node identifiers within them. And the timestamps are also independently and discretely distributed, which makes it challenging to model the historical subgraphs in a connected manner.
To achieve this, we construct a logically connected graph, named \textbf{\textit{History Temporal Graph}}, to replace the originally independent historical subgraphs, allowing a more direct approach to modeling the factual features of the previous timestamp. By utilizing connected relation features and relevant paths associated with the query, it can capture semantic connections between nodes and thus learn potential temporal linking logical rules.
% Our goal is to comprehensively model actual independent subgraphs on a logically connected graph.

Specifically, for a given query subject entity, most candidate object entities are  (directly or indirectly) connected to it in a logically connected history temporal graph. Therefore, \textbf{\textit{Temporal Path}} is proposed to comprehensively capture the connected edges between the query subject entity and other entities within the history temporal graph in a query-aware manner.
By explicitly learning the path information between entities on the history temporal graph, the query-aware representation of the history temporal path between query subject and object candidates can be obtained. Based on the learned path feature, the inference of future missing facts can be made with a final score function. A high-level idea of this approach is to induce the connection patterns and semantic correlations of the historical context.
% use query features to learn the representation of comprehensive temporal paths that are relevant to the given query.

% For a instance, given a query $(s_t,r_t,?)$ at timestamp t, we hope to focus on capture $r_t$-aware connectivity information between query subject $s_t$ and each object candidate $o_i \in \mathcal{E}$. They will be maintained at each timestamps with the topology of subgraph locally and refined with the development of time. Specifically, all relevant \emph{\textbf{relation paths}} between $s_t$ and $o_i \in \mathcal{E}$ are gathered to represent connectivity information of each pair, considering the topology of given history subgraphs $\{G_i|i<t\}$. 

We introduce the construction of history temporal graph in Section \ref{sec:graph_construct} and discuss the formulation of temporal path in Section \ref{sec:path_formulate}. Then we present the detail of query-aware temporal path processing in Section \ref{sec:path_process}. Additionally, we describe the scoring and loss function in Section \ref{sec:learning}, and finally analyze the complexity in Section \ref{sec:complexity}.

\subsection{History Temporal Graph Construction}\label{sec:graph_construct}
To comprehensively capture the connectivity patterns between entities and complex temporal characteristics in the historical subgraphs, we construct a \textit{history temporal graph}. One straightforward approach is to ignore the temporal information within the historical subgraphs and directly use all triplets from the subgraph sequence to form a complete historical graph.
However, it is essential to retain the complete information of the historical subgraph sequence. The timestamps in the historical subgraph sequence are crucial for inferring missing facts for the future. The representations of entities and relationships can vary in semantic information across different timestamps in the historical subgraphs. Additionally, there is a temporal ordering between historical subgraphs at different time periods, which further contributes to the inference process.

\begin{figure}[h]
    \centering
    \includegraphics[width=0.75\textwidth]{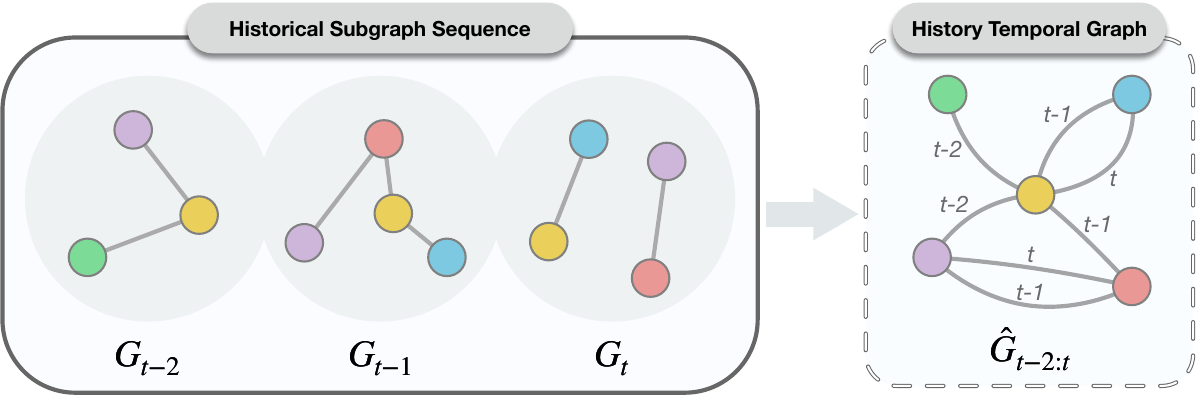}
    \caption{Example of constructing the history temporal graph $\hat{G}_{t-2:t}$ with $m=3$. For illustrative purposes, we use an undirected graph to demonstrate the construction method and omit the relationship types of edges in the historical subgraph sequence and the constructed history temporal graph. The time labels attached to the edges in the history temporal graph represent the timestamps of the corresponding edges in the historical subgraph sequence.}
    \label{fig:graph_construct}

\end{figure}

Therefore, we consider incorporating the timestamps of historical subgraphs into the history temporal graph to maximize the integrity of the original historical subgraph sequence, which allows us to retain the essential temporal context needed for inferring missing facts effectively. 
Specifically, given a query $(s,r,?,t+1)$ at future time $t+1$, we use its corresponding historical subgraph sequence $\{G_{t-m+1},..., G_t\}$ from the previous $m$ timestamps to generate the history temporal graph $\hat{G}_{t-m+1:t}$ according to the form specified in Equation \ref{eq:temporal_graph_construct}. Figure~\ref{fig:graph_construct} illustrates the construction approach of the history temporal graph.

\begin{equation}\label{eq:temporal_graph_construct}
    \hat{G}_{t-m+1:t} \leftarrow \Big\{ (s, r_\tau, o) \Big| (s,r,o)\in G_\tau , \tau \in [t-m+1,t] \Big\}
\end{equation}

Note that $r_\tau$ denotes the \textit{temporal relation} in history temporal graph $\hat{G}_{t-m+1:t}$, representing a relation type $r$ with time attribute $\tau$ attached. That is, for a temporal edge $(s, r_\tau, o) \in \hat{G}_{t-m+1:t}$, it means $(s,r,o)$ occurred in $G\tau$. For simplicity, we abbreviate the history temporal graph $\hat{G}_{t-m+1:t}$ as $\hat{G}_{\prec t+1}$ for the query at $t+1$ timestamp.
Attaching the timestamp feature to the relation can be understood as expanding the set of relation types into a combined form of \textit{relation-timestamp}, which allows us to capture the temporal aspect of relations among the entities and enables the modeling of the temporal path in history temporal graph.

\subsection{Temporal Path Formulation}\label{sec:path_formulate}
The prediction of future missing facts relies on the development trends of historical facts. To capture the linkages between the query subject entity and object candidate entities on history temporal graph, the concept of temporal path is put forward. 

\begin{definition}[]
    \label{temporalpath_define}
     \textit{\textbf{Temporal Path}} is a logical path that aggregates the semantics of all connected paths in history temporal graph between a subject entity and an object entity through arithmetic logical operations.
\end{definition}

The aggregated temporal path can be used to simultaneously represent the temporal information and comprehensive path information from the subject entity to the object entity.
This enables us to obtain a comprehensive query-aware paired representation for future facts reasoning by aggregating the information from various paths, which is from query subject entity to object candidates in history temporal graph. 

Correspondingly, in the previous work~\cite{dong2023adaptive}, DaeMon individually addresses the logical connection patterns within each independent historical subgraph. Additionally, it utilizes a path memory unit to separately model temporal path patterns along the timeline. It's crucial to note that the temporal paths learned by DaeMon are virtual since they are not acquired on a unified graph, making it impossible for practical visualization on a unified graph. In other words, the temporal path defined in DaeMon represents a latent synthesis of paths across time by the neural network, unlike the explicitly existing path on the history temporal graph in TiPNN.

It is also essential to consider the relation type information from the query while learning paired representations. That is, for a query $(s, r, ?, t+1)$, we should also embed the relation type $r$ into the paired representation (i.e., path representation) as well. This ensures that the relation type is incorporated into the learning process, enhancing the overall modeling of the query and enabling more accurate and context-aware inference of missing facts, and the detail will be discussed in Section \ref{sec:path_process}. 
Here we denote the representation of temporal path in history temporal graph $\hat{G}_{\prec t+1}$ as $\textbf{H}^{\prec t+1}_{(s,r) \rightarrow \mathcal{V}} \in \mathbb{R}^{|\mathcal{V}| \times d}$ corresponding to the query $(s,r,?)$ at future timestamp $t+1$, where $|\mathcal{V}|$ is the cardinality of object candidates set and $d$ is the dimension of temporal path embedding. Specifically, $\textbf{H}^{\prec t+1}_{(s,r) \rightarrow \mathcal{V}}$ describes the representations of temporal paths from query subject entity $s$ to all object candidate entities in $\mathcal{V}$, considering the specific query $(s,r,?)$. 
Each item in $\textbf{H}^{\prec t+1}_{(s,r) \rightarrow \mathcal{V}}$, which is denoted as $\textbf{h}^{\prec t+1}_{(s,r) \rightarrow o} \in \mathbb{R}^{d}$, represents a specific temporal path feature that learns the representation of temporal path from subject $s$ to a particular candidate object $o$, where $o \in \mathcal{V}$.

Given query $(s,r,?,t+1)$, we take an object candidate $o \in \mathcal{V}$ as an example. The temporal path between subject entity $s$ and $o$ we consider is the aggregation form of all paths that start with $s$ and end with $o$ in the topology of history temporal graph $\hat{G}_{\prec t+1}$. Formally, we describe $\textbf{h}^{\prec t+1}_{(s,r) \rightarrow o}$ as follow:

\begin{equation}\label{eq:overall_pathagg}
    \textbf{h}^{\prec t+1}_{(s,r) \rightarrow o} = 
    \bigoplus \mathcal{P}_{s \rightarrow o}^{\prec t+1} =
    \textbf{p}_1 \oplus 
    \textbf{p}_2 \oplus 
    \cdot\cdot\cdot \oplus 
    \textbf{p}_{|\mathcal{P}_{s \rightarrow o}^{\prec t+1}|}
    \Big| _{{\rm p}_k \in \mathcal{P}_{s \rightarrow o}^{\prec t+1}},
\end{equation}
where $\oplus$ denotes the paths aggregation operator that aggregates paths feature between query subject $s$ and object candidate $o$, which will be introduced in the following section; 
$\mathcal{P}_{s \rightarrow o}^{\prec t+1}$ denotes the set of paths from $s$ to $o$ in history temporal graph $\hat{G}_{\prec t+1}$, and ${|\mathcal{P}_{s \rightarrow o}^{\prec t+1}|}$ denotes the cardinality of the path set. 
A path feature $\textbf{p}_k \in \mathcal{P}_{s \rightarrow o}^{\prec t+1}$ is defined as follow when it contains edges as $(e_1 \rightarrow e_2 \rightarrow \cdot\cdot\cdot \rightarrow e_{|{\rm p}_k|})$:
\begin{equation}\label{eq:overall_pathmsg}
\textbf{p}_k^t = \textbf{w}_r(e_1) \otimes \textbf{w}_r(e_2) \otimes \cdot\cdot\cdot \otimes \textbf{w}_r(e_{|{\rm p}_k|}),
\end{equation}
where $e_{(1,2,...,|{\rm p}_k|)}$ denotes the temporal edges in path ${p}_k$, 
and $|{\rm p}_k|$ denotes the number of edges in path ${p}_k$;
$\textbf{w}_r(e_*)$ is the query-aware temporal representation of edge $e_*$ and $\otimes$ denotes the operator of merging temporal edges information within the path.

\subsection{Query-aware Temporal Path Processing}\label{sec:path_process}

\begin{figure}[t]
    \centering
    \includegraphics[width=\textwidth]{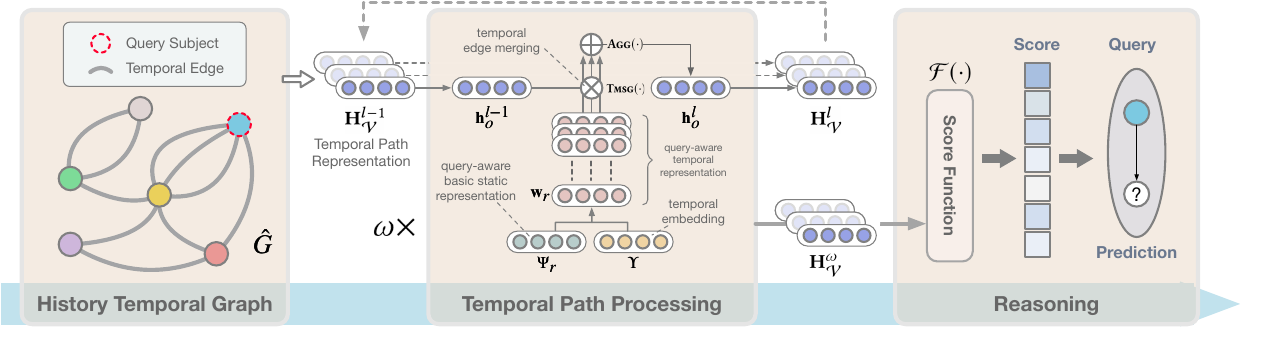}
    \caption{An illustrative diagram of the proposed TiPNN model for query-aware temporal path processing. For a given query $(s,r,?)$ at future timestamp, TiPNN engages in temporal path processing within the constructed history temporal graph $\hat{G}$ to perform prediction for the future timestamp. The temporal path feature $\textbf{H}_{\mathcal{V}}$ is iteratively learned and updated by the query-aware $\omega$-layers aggregation neural network. In each layer, the temporal edges in $\hat{G}$, enriched with temporal information, are individually modeled for basic static representation and temporal representation through $\bm{\Psi}_r$ and $\bm{\Upsilon}$, respectively. Subsequently, temporal edges are merged using $\textsc{Tmsg}(\cdot)$, followed by aggregation through $\textsc{Agg}(\cdot)$ to obtain the feature of the current layer's temporal path.}
    \label{fig:processing}

\end{figure}

In this section, we discuss how to model temporal path representations for the queries at future timestamps based on history temporal graph, as illustrated in Figure \ref{fig:processing}. It aids in predicting missing facts by providing connection patterns and temporal information into the temporal edges between entities and enabling practical inference for the future facts.

\subsubsection{Temporal Path Aggregation Layer} 

We propose a temporal path aggregation layer for comprehensively aggregating the connected temporal edges between the query subject entity $s$ and all candidate object entities in $\mathcal{V}$ within the history temporal graph.
Note that a temporal path representation is specific to a particular query. Different queries can lead to the same path, but they have different representations (e.g. when two queries share the same subject entity but have different query relations). This design ensures each path representations are context-dependent and take into account the query's unique characteristics, such as the query subject entity and query relation type, which can influence the meaning and relevance of the path in different contexts.

The entire aggregation process is carried out iteratively based on the sight outlined in Equations \ref{eq:overall_pathagg} and \ref{eq:overall_pathmsg}. 
Since the history temporal graph already contains the connection information and temporal feature of historical facts, we can directly capture the temporal path representations from history temporal graph relevant to a given query. 
We adopt the $\omega$-layers message passing approach of graph neural network at history temporal graph to expand the iterative path length and learn query-aware temporal path features, which enables the continuous and simultaneous collection of multiple path information and their corresponding temporal edge features, for the temporal path representations. 
% This approach effectively captures the temporal context and enhances the capacity to model and infer missing facts accurately for a given query.
% This approach allows us to efficiently and effectively obtain the temporal context needed for inference and prediction tasks, leveraging the comprehensive information already present in the history temporal graph.

Specifically, given a query $(s,r,?)$ at future timestamp $t+1$, based on the structural and temporal feature of historical facts, the representation of temporal path $\textbf{h}^{\prec t+1}_{(s,r) \rightarrow o}$ in history temporal graph $\hat{G}_{\prec t+1}$ (i.e., the aggregated feature of path set $\mathcal{P}_{s \rightarrow o}^{\prec t+1}$) is learned by the query-aware $\omega$-layers aggregation neural network. 
That is, $\textbf{H}^{\prec t+1}_{(s,r) \rightarrow \mathcal{V}}$ will be finally updated to represent the temporal paths from query subject $s$ to all candidate objects $o \in \mathcal{V}$ within a limited number of hops, after finishing $\omega$-th layer aggregation iteration. For the sake of simplicity, in the following text, we will use the query $(s, r, ?, t+1)$ as an example, where $s$ is the query subject entity, $r$ is the query relation type, and $t+1$ is the timestamp for the future fact to be predicted. We omit the superscript (and subscript) ${\prec t+1}$ and the indicator of query subject and relation pair $(s,r)$ in notations. $\textbf{H}^{l}_{\mathcal{V}}$ is used to denote the status of temporal path representation at the $l$-th iteration, 
and $\textbf{h}^{l}_{o} \in \textbf{H}^{l}_{\mathcal{V}}$ denotes the status of candidate $o \in \mathcal{V}$ at the $l$-th iteration, where $l \in [0,\omega]$. 
It should be stressed that $\textbf{H}^{l}_{\mathcal{V}}$ (or $\textbf{h}^{l}_{o}$) is still describing the representation of temporal path(s), rather than node(s) representation.

A temporal path is formed by logically aggregating multiple real existing paths from the history temporal graph. Each path consists of multiple temporal edges. Although the temporal edge exists in the form of \textit{relation-timestamp}, it is still based on a specific relationship $r \in \mathcal{R}$. Therefore, we initialize a trainable representation for the relation set, which serves as a foundational feature when processing temporal edge features during temporal path aggregation with query awareness.
Here, we introduce and discuss the learnable relation representation and initialization of temporal path representation.

\begin{itemize}

    \item \textbf{Learnable Relation Representation} \\ 
    At the very beginning of the processing, we initialize the representation of all relation types, which are in the set of relation types $\mathcal{R}$, with a learnable parameter $\textbf{R} \in \mathbb{R}^{|\mathcal{R}| \times d}$, where $|\mathcal{R}|$ is the cardinality of relation types set. It is essential to note that the learnable relation types representation $\textbf{R}$ is shared throughout the entire model and is not independent between each time step prediction, enabling the model to leverage the learned relationship representations consistently across different time points and queries.

    \item \textbf{Temporal Path Initialization} \\ 
    When $l=0$, $\textbf{H}^{0}_{\mathcal{V}}$ denotes the initial status of iteration that is used to prepare for the subsequent processing of temporal path iterations. Different from the common GNN-based approach and inspired by~\cite{zhu2021neural}, we initialize the temporal path feature of $o \in \mathcal{V}$ as query relation representation $\textbf{r} \in \textbf{R}$ only when $o$ is the same as the query subject entity $s$, and a zero embedding otherwise. Formally, for the query $(s,r,?)$, any candidate $o \in \mathcal{V}$ and query relation representation $\textbf{r}$, the temporal path representation $\textbf{h}^{0}_{o} \in \textbf{H}^{0}_{\mathcal{V}}$ is initialized following
    
    \begin{equation}
        \textbf{h}^{0}_{o}\leftarrow\begin{cases}
        \textbf{r}  &  if\ o \Leftrightarrow s,\\
        \vec 0      &  if\ o \nLeftrightarrow s.
        \end{cases}
    \end{equation}
    
    This initialization strategy ensures that the initial state is contextually sensitive to the given query, starting at the query subject with the query relation feature, providing a foundation for subsequent iterations to capture and model the temporal paths.
    
\end{itemize}

With the initialization of temporal path representations, now we introduce the iterative aggregation process of the temporal path. Taking $\textbf{h}_o^l$ as an example,

\begin{equation}\label{eq:aggandmsg}
\textbf{h}_o^l = \textsc{Agg} \bigg(
\Big\{
\textsc{Tmsg}\big(\textbf{h}_{z}^{l-1},\textbf{w}_r(z,p_\tau,o)\big) 
\Big|(z,p_\tau,o)\in \hat{G}_{\prec t+1}
\Big\}
\bigg),
\end{equation}
where $\textsc{Agg}(\cdot)$ and $\textsc{Tmsg}(\cdot)$ are the aggregation and temporal edge merging function, corresponding to the operator in Equation \ref{eq:overall_pathagg} and \ref{eq:overall_pathmsg}, respectively, which will be introduced in the following section; 
$\textbf{w}_r \in \mathbb{R}^{d}$ denotes query-aware temporal representation of a temporal edge type; 
$(z,p_\tau,o)\in \hat{G}_{\prec t+1} $ is a temporal edge in history temporal graph $\hat{G}_{\prec t+1}$ that temporal relation $p_\tau$ occurs between an entity $z$ and candidate entity $o$ at the history time $\tau$. 

However, to emphasize the distinction from DaeMon~\cite{dong2023adaptive}, it is important to note that in DaeMon, the acquisition of temporal paths is learned independently on each isolated historical subgraph (i.e., $G_i$, where $i < t+1$). Consequently, the final temporal path representation requires stacking $m$-times (i.e., history length) updates of $\textbf{h}_o$ in DaeMon.

% Equation \ref{eq:aggandmsg} describes the iterative aggregation process of $\textbf{h}_o$ in the $l$-th layer. 
% Intuitively, the $\textsc{Tmsg}(\cdot)$ function is responsible for iteratively searching for paths with a candidate object entity $o$ as the endpoint, continuously propagating information along the temporal edges. 
% On the other hand, the $\textsc{Agg}(\cdot)$ function is responsible for aggregating the information obtained from each iteration along the temporal edges. 
% Due to the initial state of $\textbf{H}_o^0$, where only $\textbf{h}_s^0$ is given the initial information, after $\omega$ iterations of aggregation, $\textbf{h}_o^\omega$ captures the comprehensive semantic connection and temporal dependency information of multiple paths from subject node $s$ to object candidate $o$ in the history temporal graph $\hat{G}_{\prec t+1}$.

\subsubsection{Aggregation Operating} \label{sec:operating}

Equation \ref{eq:aggandmsg} describes the iterative aggregation process of $\textbf{h}_o$ in the $l$-th layer. Here we provide explanations for the temporal edge merging function $\textsc{Tmsg}(\cdot)$ and path aggregation function $\textsc{Agg}(\cdot)$, respectively.
Intuitively, the $\textsc{Tmsg}(\cdot)$ function is responsible for iteratively searching for paths with a candidate object entity $o$ as the endpoint, continuously propagating information along the temporal edges. 
On the other hand, the $\textsc{Agg}(\cdot)$ function is responsible for aggregating the information obtained from each iteration along the temporal edges. 
Due to the initial state of $\textbf{H}_o^0$, where only $\textbf{h}_s^0$ is given the initial information, after $\omega$ iterations of aggregation, $\textbf{h}_o^\omega$ captures the comprehensive semantic connection and temporal dependency information of multiple paths from subject $s$ to object candidate $o$ in the history temporal graph $\hat{G}_{\prec t+1}$.

\begin{itemize}
    \item \textbf{Temporal Edge Merging Function $\textsc{Tmsg}(\cdot)$}    \\
    It takes the current temporal path representation $\textbf{h}_{z}^{l-1}$, query relation representation $\textbf{r}$, and the temporal edge information as input and calculates the updated potential path information for the candidate object $o$. 
    Similar to message passing in knowledge graphs, we have adopted a vectorized multiplication method \cite{yang2014embedding} to achieve feature propagation in history temporal graph, following

    \begin{equation}\label{eq:tmsg}
        % \textsc{Tmsg}\big(\textbf{h}_{z}^{t,l-1},\textbf{w}_r(z,p,o_i)\big) = 
        \textsc{Tmsg}\big(\textbf{h}_{z}^{l-1},\textbf{w}_r(z,p_\tau,o)\big) = 
        \textbf{h}_{z}^{l-1} \otimes \textbf{w}_r(z,p_\tau,o),
    \end{equation}

    where the operator $\otimes$ is defined as element-wise multiplication between $\textbf{h}_{z}^{l-1}$ and $\textbf{w}_r(z,p_\tau,o)$. The vectorized multiplication can be understood as scaling $\textbf{h}_{z}^{l-1}$ by $\textbf{w}_r(z,p_\tau,o)$ in our temporal edges merging~\cite{yang2014embedding}. $\textbf{w}_r(z,p_\tau,o)$ represents the query-aware temporal representation of a temporal edge $(z,p_\tau,o)$, which needs to ensure its relevance to the query. 
    Different from DaeMon~\cite{dong2023adaptive}, in the history temporal graph, each edge is appended with temporal information, preventing the consideration of features solely based on static relation types.
    Therefore, the approach of {temporal relation encoding} is proposed for the temporal edge representation.
    
    % \begin{itemize}
        % \item \textbf{\textit{Temporal Relation Encoding}} \\
        \textit{Temporal Relation Encoding.}
        As mentioned earlier, temporal edge representation should align with the query's characteristics. We take into account the temporal edge features from the query relation representation, semantic and temporal characteristics of the temporal edges in the history temporal graph, generating comprehensive embedding with both static and temporal characteristics. Formally, $\textbf{w}_r(z,p_\tau,o)$ can be derived as 

        \begin{equation}
            \textbf{w}_r(z,p_\tau,o) = 
            g\Bigg(
            \bm{\Psi}_r(p) 
            \bigg|\bigg| 
            \bm{\Upsilon}(\Delta\tau)\Bigg),
        \end{equation}

        where $\bm{\Psi}_r(p) \in \mathbb{R}^{d}$ denotes the query relation $r$-aware basic static representation of edge type $p$, 
        $\bm{\Upsilon}(\Delta\tau) \in \mathbb{R}^{d}$ denotes the temporal embedding of the temporal edge type $p_\tau$,
        $||$ denotes the operator of concatenation,
        and $g(\cdot)$ is a feed-forward neural network. 
        For the basic static representation of $p$, following~\cite{dong2023adaptive}, we obtain it through a linear transformation. Based on the query relation representation $\mathbf{r}$, we map a representation as the static semantic information of the temporal edge type $p$, following

        \begin{equation}
            \bm{\Psi}_r(p) = 
            \textbf{W}_p \textbf{r}+\textbf{b}_p,
        \end{equation}

        where $\textbf{W}_p$ and $\textbf{b}_p$ are learnable parameters and serve as the weights and biases of the linear transformation, respectively, 
        which makes the basic static information of the temporal edges derived from the given query relation embedding, ensuring the awareness of the query.
        And for the temporal embedding of the temporal edge, we use generic time encoding \cite{xu2020inductive} to model the temporal information in temporal edges, following

        \begin{equation}
            \Delta\tau = | \tau - t_q |,
        \end{equation}

        \begin{equation}\label{eq:time_encode}
            \bm{\Upsilon}(\Delta\tau) = 
            \sqrt{\frac{1}{d}} \ \  
            \Big[ cos(\bm{w}_1\Delta\tau+\bm\phi_1), cos(\bm{w}_2\Delta\tau+\bm\phi_2), \cdots, 
            cos(\bm{w}_d\Delta\tau+\bm\phi_d) \ \Big],
        \end{equation}

        where $t_q$ denotes the query timestamp, $\Delta\tau$ denotes the time interval between the query timestamp $t_q$ and the temporal edge timestamp $\tau$, which measures how far apart $\tau$ is from $t_q$, namely \textit{relative time distance}. Moreover, $\bm{w}_*$ and $\bm\phi_*$ are learnable parameters, and $d$ is the dimension of the vector representation, which is the same as the dimension of the static representation.
    % \end{itemize}

    The $\textsc{Tmsg}(\cdot)$ function effectively bridges the temporal dependencies and semantic connections between the query subject and the candidate object through propagation along the temporal edges in paths.
    Besides, in the encoder of temporal relation, 
    considering query relation feature allows to tailor the temporal edge representation to the specific relation type relevant to the query; 
    temporal characteristics of edges help to model chronological order and time-based dependencies between different facts;
    and simultaneously modeling the basic static and temporal features of the temporal edge make more comprehensive semantics obtained.
    % We consider several factors to model the temporal edges in history temporal graph. These factors include:
    % Query relationship features: We incorporate information about the relationship type from the given query. This allows us to tailor the temporal edge representation to the specific relationship type relevant to the query.
    % Semantic characteristics of edges in the history temporal graph: We take into account the semantic information of the edges, which captures the meaning and context of the relationships between entities at different time points.
    % Temporal characteristics of edges in the history temporal graph: We also consider the temporal information of the edges, which captures the chronological order and time-based dependencies between different factual events.
    
    \item \textbf{Path Aggregation Function $\textsc{Agg}(\cdot)$}  \\
    It considers the accumulated information from the previous iteration and the newly propagated information along the temporal edge to update the temporal path representation for a candidate object. We adopt principal neighborhood aggregation (PNA) proposed in~\cite{corso2020principal}, which leverages multiple aggregators (namely mean, maximum, minimum, and standard deviation) to learn joint feature, since previous work has verified its effectiveness~\cite{zhu2022neural,pmlr-v162-miao22a}. We also consider the traditional aggregation function as a comparison, such as sum, mean, and max, which will be introduced in Section \ref{sec:ablation}.
    
\end{itemize}

Finally, after the $\omega$-th iteration of aggregation, $\textbf{H}_{\mathcal{V}}^\omega$ will obtain the representation of the temporal paths from query subject $s$ to all candidate objects in $\mathcal{V}$.
Here we should note that, unlike the previous work \cite{dong2023adaptive}, TiPNN integrates all historical subgraphs into a unified history temporal graph $\hat{G}$. When handling temporal path features, any temporal edge $(z,p_\tau,o)\in \hat{G}$ is associated with timestamp information. Additionally, the static representation of $p_\tau$ and its temporal representation are respectively processed by $\bm{\Psi}$ and $\bm{\Upsilon}$. Consequently, TiPNN eliminates the need, as in DaeMon, for a separate consideration of cross-temporal path fusion. The resulting $\textbf{H}_{\mathcal{V}}^\omega$ already encompasses comprehensive semantic connections and temporal dependency information from multiple paths in history.
% with comprehensive semantic connection and temporal dependency information of multiple paths.

\subsection{Learning and Inference}\label{sec:learning} 

TiPNN models the query-aware temporal path feature by aggregation process within the history temporal graph. By capturing comprehensive embedding with both static and temporal characteristics of temporal edges, and aggregating multiple paths that consist of temporal edges, TiPNN learns a comprehensive representation of temporal path. Different from most previous models, we utilize the temporal edge features in a path from query subject to candidate object, without considering any entity embedding during modeling processing, thus TiPNN can solve the inductive setting, which will be presented in Section \ref{sec:exp_result} in detail.

\subsubsection{Score Function}\label{sec:score_func}

Here we show how to apply the final learned temporal paths representation to the temporal knowledge graph reasoning.
Given query subject $s$ and query relation $r$ at timestamp $t+1$, after obtaining temporal paths representation $\textbf{H}_{(s,r)\rightarrow {\mathcal{V}}}^{\omega}$,
we predict the conditional likelihood of the future object candidate $o \in \mathcal{V}$ using $\textbf{h}_{(s,r)\rightarrow {o}}^{\omega} \in \textbf{H}_{(s,r)\rightarrow {\mathcal{V}}}^{\omega}$, following:

\begin{equation}
    p(o|s,r)=
    \sigma\Bigg(\mathcal{F}\Big(
    \textbf{h}_{(s,r)\rightarrow {o}}^{\omega}\ \big|\big|\ \bm{r}
    \Big)
    \Bigg),
\end{equation}
where $\mathcal{F(\cdot)}$ is a feed-forward neural network, $\sigma(\cdot)$ is the sigmoid function and $||$ denotes embedding concatenation. Note that we append the query relation embedding $\bm r$ to the temporal path feature $\textbf{h}_{(s,r)\rightarrow {o}}^{\omega}$, and it helps to alleviate the insensitivity to unreachable distances for nodes within a limited number of hops, and also enhances the learning capacity of relation embeddings $\textbf{R}$.

As we have added inverse quadruple $(o,r^{-1},s,t)$ corrsponding to $(s,r,o,t)$ into the dataset in advance, without loss of generality, we can also predict subject $s \in \mathcal{V}$ given query relation $r^{-1}$ and query object $o$ with the same model as:

\begin{equation}
    p(s|o,r^{-1})=
    \sigma\Bigg(\mathcal{F}\Big(
    \textbf{h}_{(o,r^{-1})\rightarrow {s}}^{\omega}\ \big|\big|\ \bm{r}^{-1}
    \Big)
    \Bigg).
\end{equation}

\subsubsection{Parameter Learning}

Reasoning on a given query can be seen as a binary classification problem. 
The objective is to minimize the negative log-likelihood of positive and negative triplets, as shown in Equation \ref{eq:tkg_loss}.
In the process of generating negative samples, we follow the Partial Completeness Assumption ~\cite{galarraga2013amie}. Accordingly, for each positive triplet in the reasoning future triplet set, we create a corresponding negative triplet by randomly replacing one of the entities with a different entity. It ensures that the negative samples are derived from the future triplet set at the same timestamp, allowing the model to effectively learn to discriminate between correct and incorrect predictions at the same future timestamp. 

\begin{equation}\label{eq:tkg_loss}
\mathcal{L}_{TKG} = 
-\log p(s,r,o)- \sum_{j=1}^{n} \frac{1}{n} log(1-p(\overline{s_j},r,\overline{o_j})),
\end{equation}
where $n$ is hyperparameter of negative samples number per positive sample; $(s,r,o)$ and $(\overline{s_j},r,\overline{o_j})$ are the positive sample and $j$-th negative sample, respectively.

Besides, to promote orthogonality in the learnable raltion parameter $\textbf{R}$ initialized at the beginning, a regularization term is introduced in the objective function, inspired by the work in \cite{xu2020variational}. The regularization term is represented following Equation \ref{eq:reg_loss}, where $\textbf{I}$ is the identity matrix, $\alpha$ is a hyperparameter and $\|\cdot\|$ denotes the L2-norm.

\begin{equation}\label{eq:reg_loss}
\mathcal{L}_{REG} = \Big\| \ \textbf{R}^T\textbf{R}-\alpha\textbf{I} \ \Big\|
\end{equation}

Therefore, the final loss of TiPNN is the sum of two losses and can be denoted as:

\begin{equation}
    \mathcal{L}=\mathcal{L}_{TKG}+\mathcal{L}_{REG}.
\end{equation}

\subsection{Complexity Analysis}\label{sec:complexity}

The score of each temporal path from the query subject and candidate object can be calculated in parallel operation. Therefore, to see the complexity of the proposed TiPNN, we analyze the computational complexity of each query.
The major operation in TiPNN is query-aware temporal path processing, and each aggregation iteration contains two steps: temporal edge merging function $\textsc{Tmsg}(\cdot)$ and path aggregation function $\textsc{Agg}(\cdot)$ (as shown in Equation \ref{eq:aggandmsg}).
We use $|\mathcal{E}|$ to denote the maximum number of concurrent facts in the historical subgraph sequence, and $|\mathcal{V}|$ to denote the cardinality of the entity set.
$\textsc{Tmsg}(\cdot)$ has the time complexity of $O(m|\mathcal{E}|)$ since it performs Equation \ref{eq:tmsg} for all edges in the history temporal graph, where $m$ denotes the length of the history used for reasoning.
$\textsc{Agg}(\cdot)$ has the time complexity of $O(|\mathcal{V}|)$ since the aggregation method we adopt will perform for every entity. 
And $\textsc{Agg}(\cdot)$ is executed after $\textsc{Tmsg}(\cdot)$ at each aggregation iteration, so the time complexity is finally $O(\omega(m|\mathcal{E}|+|\mathcal{V}|))$ when the number of aggregation layers is $\omega$.

\section{Experiments}

In this section, we carry out a series of experiments to assess the effectiveness and performance of proposed TiPNN. Through these experiments, we aim to address and answer the following research questions:

\begin{itemize}
    \item \textbf{RQ1:} How does the proposed model perform as compared with state-of-the-art knowledge graph reasoning and temporal knowledge graph reasoning methods?
    \item \textbf{RQ2:} How does each component of TiPNN (i.e., temporal relation encoding, temporal edge merging function, path aggregation function) affect the performance?
    \item \textbf{RQ3:} How do the core parameters in history temporal graph construction  (i.e., sampling history length) and query-aware temporal path processing (i.e., number of temporal path aggregation layers) affect the reasoning performance?
    \item \textbf{RQ4:} How does the performance efficiency of TiPNN compared with state-of-the-art method for the reasoning?
    \item \textbf{RQ5:} How does the proposed model solve the inductive setting on temporal knowledge graph reasoning?
    \item \textbf{RQ6:} How does the proposed model provide reasoning evidence based on the history temporal path for the reasoning task?
\end{itemize}

\subsection{Experimental Settings}
\subsubsection{Datasets}

We conduct extensive experiments on four representative temporal knowledge graph datasets, namely, ICEWS18~\cite{jin2020Renet}, GDELT~\cite{2013GDELT}, WIKI~\cite{2018Deriving}, and YAGO~\cite{2013YAGO3}. These datasets are widely used in the field of temporal knowledge reasoning due to their diverse temporal information.
The ICEWS18 dataset is collected from the Integrated Crisis Early Warning System~\cite{DVN/28075_2015}, which records various events and activities related to global conflicts and crises, offering valuable insights into temporal patterns of such events. The GDELT dataset~\cite{jin2019recurrent} is derived from the Global Database of Events, Language, and Tone, encompassing a broad range of events across the world and enabling a comprehensive analysis of global events over time. 
WIKI~\cite{2018Deriving} and YAGO~\cite{2013YAGO3} are two prominent knowledge bases containing factual information with explicit temporal details. WIKI covers a wide array of real-world knowledge with timestamps, while YAGO offers a structured knowledge base with rich semantic relationships and temporal annotations. And we focus on the subsets of WIKI and YAGO that have yearly granularity in our experiments.
The details of the datasets are provided in Table \ref{tab:data_statistics}.   
To ensure fair evaluation, we adopt the dataset splitting strategy employed in ~\cite{jin2020Renet}. The dataset is partitioned into three sets: training, validation, and test sets, based on timestamps. Specifically, (timestamps of the train) $<$ (timestamps of the valid) $<$ (timestamps of the test).

\begin{table}[!htbp]
    \centering
    % \small
    % \scriptsize
    % \footnotesize
    \setlength\tabcolsep{10pt}  
    \caption{Statistics of Datasets ($\mathcal{E}_{train}$, $\mathcal{E}_{valid}$, $\mathcal{E}_{test}$ are the numbers of facts in training, validation, and test sets.).}
    \begin{tabular}{ccccccc}
    \toprule
    % \hline
       Datasets  & $|\mathcal{V}|$ & $|\mathcal{R}|$ & $\mathcal{E}_{train}$ & $\mathcal{E}_{valid}$ & $\mathcal{E}_{test}$ & $|\mathcal{T}|$\\
    \midrule
    ICEWS18 & 23,033 & 256 & 373,018 & 45,995 &  49,545 & 304 \\
    GDELT & 7,691 & 240 & 1,734,399 &  238,765 & 305,241 & 2,976 \\
    WIKI & 12,554 & 24 & 539,286 & 67,538 & 63,110 & 232 \\
    YAGO & 10,623 & 10 & 161,540 & 19,523 & 20,026 & 189 \\
    \bottomrule
    \end{tabular}
    
    \label{tab:data_statistics}
    % \vspace{-4mm}
\end{table}

\subsubsection{Evaluation Metrics}

To evaluate the performance of the proposed model for TKG reasoning, we employ the commonly used task of link prediction on future timestamps. 
To measure the performance of the method, we report the Mean Reciprocal Rank (MRR) and Hits@\{1, 3, 10\} metrics. MRR is a measure that evaluates the average reciprocal rank of correctly predicted facts for each query. Hits@{k} is a set of metrics that measures the proportion of queries for which the correct answer appears in the top-k positions of the ranking list.

In contrast to the traditional filtered setting used in previous works~\cite{bordes2013translating,jin2020Renet,zhu2021learning}, where all valid quadruples appearing in the training, validation, or test sets are removed from the ranking list of corrupted facts, we consider that this setting is not suitable for TKG reasoning tasks. 
Instead, we opt for a more reasonable evaluation, namely time-aware filtered setting, where only the facts occurring at the same time as the query are filtered from the ranking list of corrupted facts, which is aligned with recent works~\cite{2021TimeTraveler,han2021learning,dong2023adaptive}.

\subsubsection{Compared Methods}

We conduct a comprehensive comparison of our proposed model with three categories of baselines:
(1) \textit{{KG Reasoning Models.}} 
It consists of traditional KG reasoning models that do not consider timestamps, including DistMult\cite{yang2014embedding}, ComplEx~\cite{2016Complex}, ConvE~\cite{2017Convolutional}, and RotatE~\cite{2019RotatE}. 
(2) \textit{{Interpolated TKG Reasoning models.}} 
We also compare our proposed model with four interpolated TKG reasoning methods, namely TTransE~\cite{2018Deriving}, TA-DistMult~\cite{A2018Learning}, DE-SimplE~\cite{2020Diachronic}, and TNTComplEx~\cite{2020Tensor}. 
(3) \textit{{Extrapolated TKG Reasoning models.}}
We include state-of-the-art extrapolated TKG reasoning methods that infers future missing facts, including TANGO-Tucker~\cite{han2021learning}, TANGO-DistMult~\cite{han2021learning}, CyGNet~\cite{zhu2021learning}, RE-NET~\cite{jin2020Renet}, RE-GCN~\cite{li2021temporal}, TITer~\cite{sun2021timetraveler}, xERTE~\cite{han2020explainable}, CEN~\cite{li-etal-2022-complex}, GHT~\cite{sun2022graph}, and DaeMon~\cite{dong2023adaptive}.

\subsubsection{Implementation Details}

For history temporal graph construction, we perform a grid search on the history length $m$ and present overview results with the lengths 25, 15, 10, 8, corresponding to the datasets ICEWS18, GDELT, WIKI, and YAGO in Table \ref{tab:overall_results}, which is described in detail presented in Figure \ref{fig:history_length}. 
The embedding dimension $d$ is set to 64 for temporal path representation.
For query-aware temporal path processing, we set the number of temporal path aggregation layers $\omega$ to 6 for ICEWS18 and GDELT datasets, and 4 for WIKI and YAGO datasets.
We conduct layer normalization and shortcut on the aggregation layers. The activation function $relu$ is adopted for the aggregation of the temporal path.  Similar to \cite{zhu2021neural}, we use different temporal edge representations in different aggregation layers, that is the learnable parameter in $\bm{\Psi}$ and $\bm{\Upsilon}$ is independent of each aggregation iteration \cite{10.5555/3454287.3455662} \cite{yang2017differentiable}.
To facilitate parameter learning, we have set the number of negative sampling to 64 and the hyperparameter $\alpha$ in the regularization term to 1. We use Adam~\cite{kingma2014adam} for parameter learning, with a learning rate of ${1e-4}$ for YAGO and ${5e-4}$ for others. The maximum epoch of training has been set to 20.
We also design an experiment in an inductive setting, which predicts facts with unseen entities in the training set, to demonstrate the inductive ability of our proposed model. All experiments were conducted with EPYC 7742 CPU and 8 TESLA A100 GPUs.
% In addition, we have released multi-device parallel version code to accelerate the training and inference. Codes and datasets are all available at https://github.com/hhdo/DaeMon. 

\subsection{Experimental Results (RQ1)}\label{sec:exp_result}
\begin{table*}[!htbp]
    \centering
    % \small
    % \scriptsize
    \footnotesize
    \setlength\tabcolsep{1.85pt}
    \caption{Overall Performance Comparison of Different Methods. Evaluation metrics are time-aware filtered MRR and Hits@\{1,3,10\}. All results are multiplied by 100. The best results are highlighted in \textbf{bold}. And the second-best results are highlighted in \underline{underline}. (Higher values indicate better performance.)}
    \begin{tabular}{c|cccc|cccc|cccc|cccc}

    \toprule
    \multirow{2.7}{*}{Model} &  \multicolumn{4}{c|}{ICEWS18} & \multicolumn{4}{c|}{GDELT} & \multicolumn{4}{c|}{WIKI} & \multicolumn{4}{c}{YAGO} \\
    \cmidrule{2-17}
    & MRR & H@1 & H@3 & H@10 & MRR & H@1 & H@3 & H@10 & MRR & H@1 & H@3 & H@10 & MRR & H@1 & H@3 & H@10 \\ 
    \midrule
    DistMult & 11.51 & 7.03 & 12.87 & 20.86 & 8.68 & 5.58 & 9.96 & 17.13 & 10.89 & 8.92 & 10.97 & 16.82 & 44.32 & 25.56 & 48.37 & 58.88 \\ 
    ComplEx & 22.94 & 15.19 & 27.05 & 42.11 & 16.96 & 11.25 & 19.52 & 32.35 & 24.47 & 19.69 & 27.28 & 34.83 & 44.38 & 25.78 & 48.20 & 59.01 \\ 
    ConvE & 24.51 & 16.23 & 29.25 & 44.51 & 16.55 & 11.02 & 18.88 & 31.60 & 14.52 & 11.44 & 16.36 & 22.36 & 42.16 & 23.27 & 46.15 & 60.76 \\ 
    RotatE & 12.78 & 4.01 & 14.89 & 31.91 & 13.45 & 6.95 & 14.09 & 25.99 & 46.10 & 41.89 & 49.65 & 51.89 & 41.28 & 22.19 & 45.33 & 58.39 \\ 
    \midrule
    TTransE & 8.31 & 1.92 & 8.56 & 21.89 & 5.50 & 0.47 & 4.94 & 15.25 & 29.27 & 21.67 & 34.43 & 42.39 & 31.19 & 18.12 & 40.91 & 51.21  \\ 
    TA-DistMult & 16.75 & 8.61 & 18.41 & 33.59 & 12.00 & 5.76 & 12.94 & 23.54 & 44.53 & 39.92 & 48.73 & 51.71 & 54.92 & 48.15 & 59.61 & 66.71  \\ 
    DE-SimplE & 19.30 & 11.53 & 21.86 & 34.80 & {19.70} & 12.22 & {21.39} & 33.70 & 45.43 & 42.60 & 47.71 & 49.55 & 54.91 & 51.64 & 57.30 & 60.17  \\ 
    TNTComplEx & 21.23 & 13.28 & 24.02 & 36.91 & 19.53 & 12.41 & 20.75 & 33.42 & 45.03 & 40.04 & 49.31 & 52.03 & 57.98 & 52.92 & 61.33 & 66.69 \\ 
    \midrule
    TANGO-Tucker & 28.68 & 19.35 & 32.17 & 47.04 & 19.42 & 12.34 & 20.70 & 33.16 & 50.43 & 48.52 & 51.47 & 53.58 & 57.83 & 53.05 & 60.78 & 65.85  \\ 
    TANGO-DistMult & 26.65 & 17.92 & 30.08 & 44.09 & 19.20 & 12.17 & 20.40 & 32.78 & 51.15 & 49.66 & 52.16 & 53.35 & 62.70 & 59.18 & 60.31 & 67.90  \\ 
    CyGNet & 24.93 & 15.90 & 28.28 & 42.61 & 18.48 & 11.52 & 19.57 & 31.98 & 33.89 & 29.06 & 36.10 & 41.86 & 52.07 & 45.36 & 56.12 & 63.77  \\ 
    RE-NET & 28.81 & 19.05 & 32.44 & 47.51 & 19.62 & {12.42} & 21.00 & {34.01} & 49.66 & 46.88 & 51.19 & 53.48 & 58.02 & 53.06 & 61.08 & 66.29  \\ 
    RE-GCN & {30.58} & 21.01 & 34.34 & {48.75} & 19.64 & 12.42 & 20.90 & 33.69 & {77.55} & {73.75} & {80.38} & {83.68} & 84.12 & 80.76 & 86.30 & 89.98  \\ 
    TITer & 29.98 & {22.05} & 33.46 & 44.83 & 15.46 & 10.98 & 15.61 & 24.31 & 75.50 & 72.96 & 77.49 & 79.02 & {87.47} & {84.89} & {89.96} & {90.27}  \\ 
    xERTE & 29.31 & 21.03 & {33.40} & 45.60 & 18.09 & 12.30 & 20.06 & 30.34 & 71.14 & 68.05 & 76.11 & 79.01 & 84.19 & 80.09 & 88.02 & 89.78 \\ 
    CEN & 30.84 & 21.23 & 34.58 & 49.67 & 20.18 & 12.84 & 21.51 & 34.10 & 78.35 & 74.69 & 81.47 & 84.45 & 83.49 & 79.77 & 85.85 & 89.92 \\ 
    GHT & 29.16 & 18.99 & 33.16 & 48.37 & 20.13 & 12.87 & 21.30 & 34.19 & 48.50 & 45.08 & 50.87 & 53.69 & 57.22 & 51.64 & 60.68 & 67.17 \\ 
    DaeMon & {\ul 31.85} & {\ul 22.67} & {\ul 35.92} & {\ul 49.80} & {\ul 20.73} & {\ul 13.65} & {\ul 22.53} & {\ul 34.23} & {\ul 82.38} & {\ul 78.26} & {\ul 86.03} & {\ul 88.01} & {\ul 91.59} & {\ul 90.03} & {\ul 93.00} & {\ul 93.34} \\ 
    
    \midrule
    \textbf{TiPNN} & \textbf{32.17} & \textbf{22.74} & \textbf{36.24} & \textbf{50.72} & \textbf{21.17} & \textbf{14.03} & \textbf{22.98} & \textbf{34.76} & \textbf{83.04} & \textbf{79.04} & \textbf{86.45} & \textbf{88.54} & \textbf{92.06} & \textbf{90.79} & \textbf{93.15} & \textbf{93.58} \\

    \bottomrule
    \end{tabular}
    
    \label{tab:overall_results}

\end{table*}

The experiment results on the TKG reasoning task are shown in Table \ref{tab:overall_results} in terms of time-aware filtered MRR and Hits@\{1,3,10\}. 
The results demonstrate the effectiveness of our proposed model with a convincing performance and validate the superiority of TiPNN in handling temporal knowledge graph reasoning tasks. It consistently outperforms all baseline methods and achieves state-of-the-art (SOTA) performance on the four TKG datasets.

In particular, TiPNN exhibits better performance compared to all static models (listed in the first block of Table \ref{tab:overall_results}) and the temporal models in the interpolation setting (presented in the second block of Table \ref{tab:overall_results}). This is attributed to TiPNN's incorporation of temporal features of facts and its ability to learn temporal inductive paths, which enables it to effectively infer future missing facts. By leveraging comprehensive historical semantic information, TiPNN demonstrates remarkable proficiency in handling temporal knowledge graph reasoning tasks.
Compared with the temporal models under the extrapolation setting (those presented in the third block of Table \ref{tab:overall_results}), the proposed model also achieves better results. TiPNN demonstrates its capability of achieving superior results by effectively modeling temporal edges. In contrast to the previous model, it starts from the query and obtains query-specific temporal path representations, allowing for more accurate predictions about the future. Thanks to the message propagation mechanism between temporal edges, it leverages both temporal and structural information to learn customized representations tailored to each query. This enables TiPNN to make more precise predictions for future missing facts. The impressive performance of DaeMon demonstrated its ability to model relation features effectively. However, we further enhance the integration of relation and temporal features by weakening the barriers between different time-stamped subgraphs using the constructed history temporal graph. It enables us to capture the characteristics of historical facts over a larger span of information and reduces the parameter loss caused by subgraph evolution patterns.

\subsection{Ablation Study (RQ2)}\label{sec:ablation}
Note that TiPNN involves three main operational modules when processing query-aware temporal paths: the temporal relation encoder, the temporal edge merging operation, and the path aggregation operation. The temporal relation encoder is responsible for learning the features of temporal edges. The temporal edge merging operation is used to generate path features for each path from the query subject entity to the candidate object entity. The path aggregation module is responsible for consolidating all path features for pair-wise representations of temporal paths.

To show the impact of these components, we conducted an ablation analysis on the temporal encoder $\bm\Upsilon$ of temporal relation encoding and discussed the influence of temporal encoding independence on the results. We also compared three variants of the temporal edge merging operation in $\textsc{Tmsg}(\cdot)$, as well as replacements for the aggregator in the path aggregation module $\textsc{Agg}(\cdot)$. These comprehensive evaluations allowed us to assess the effectiveness of each component and gain insights into their respective contributions to the overall performance of TiPNN.

\subsubsection{Effectiveness of Temporal Relation Encoding}

\begin{figure}[!htbp]
\centering
\subfigure[ICEWS18]{
\includegraphics[width=6.5cm]{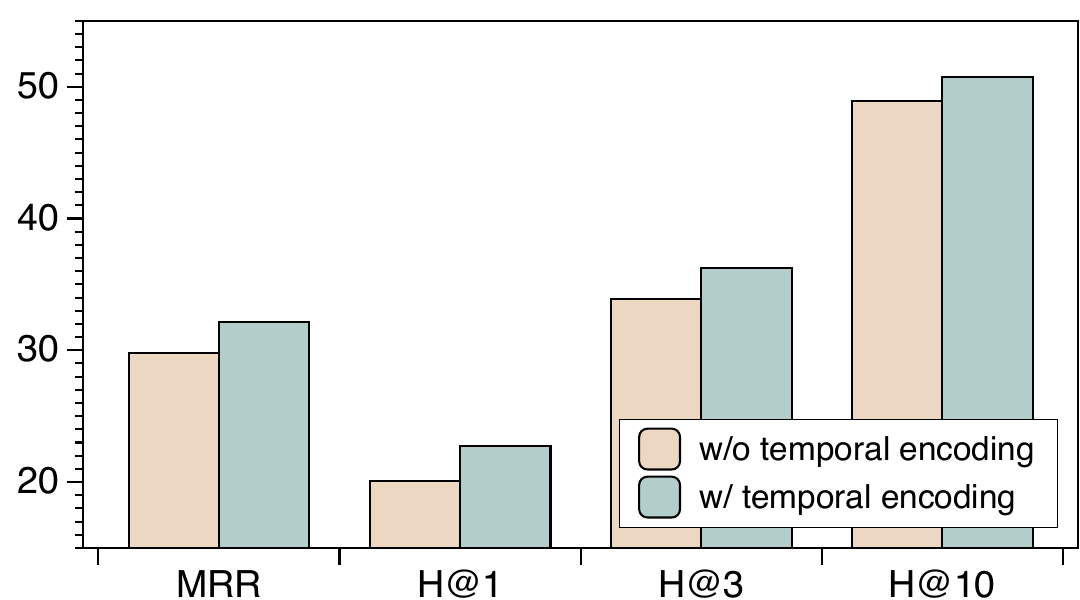}
}
% \hspace{5mm}
% \vspace{-3mm}
\subfigure[GDELT]{
\includegraphics[width=6.5cm]{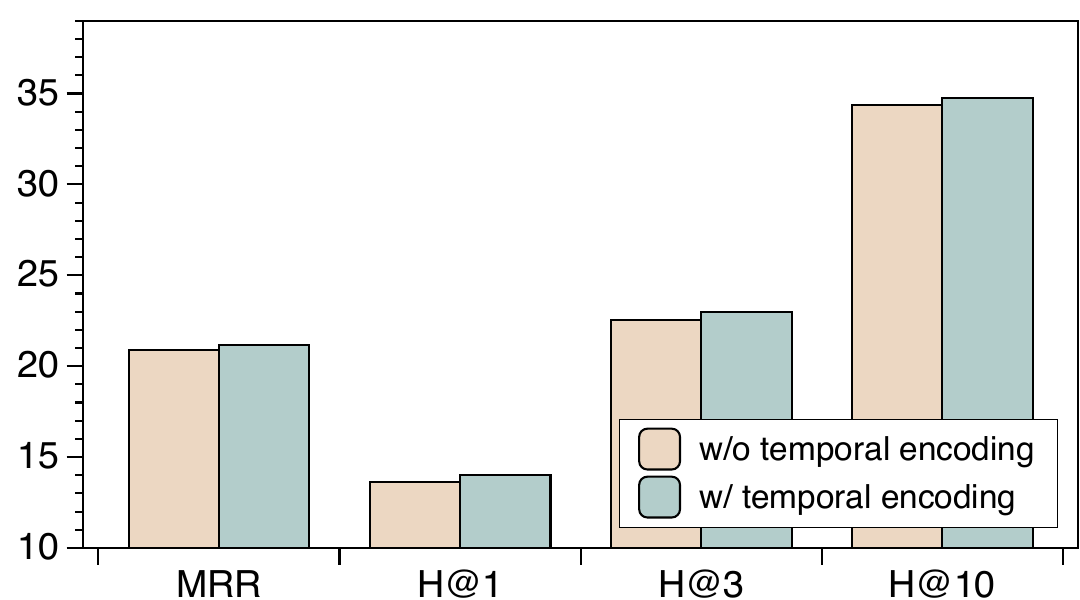}
}
\\
\subfigure[WIKI]{
\includegraphics[width=6.5cm]{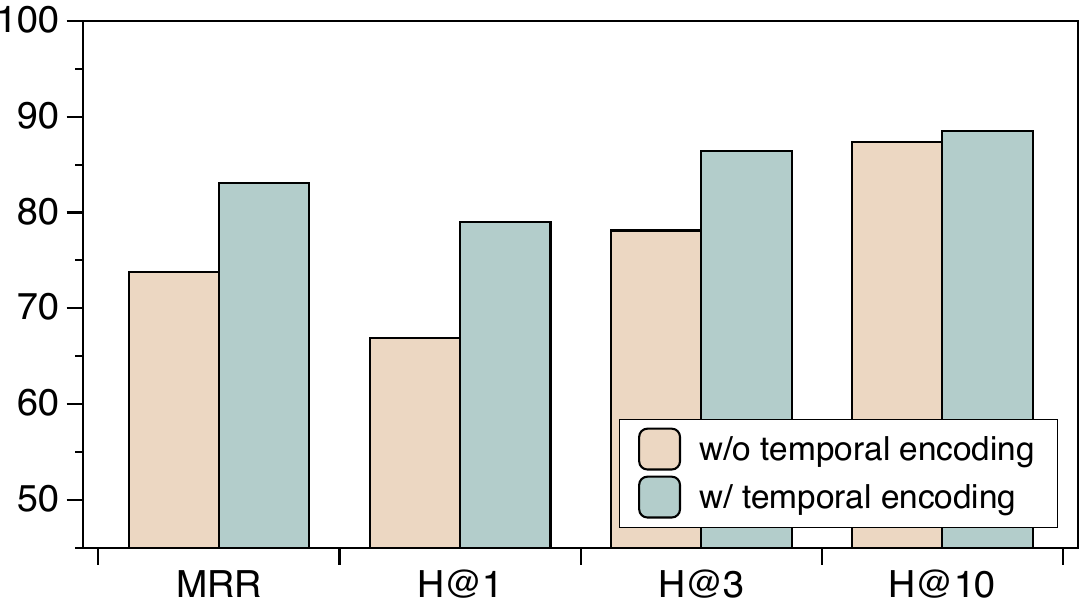}
}
% \hspace{5mm}
\subfigure[YAGO]{
\includegraphics[width=6.5cm]{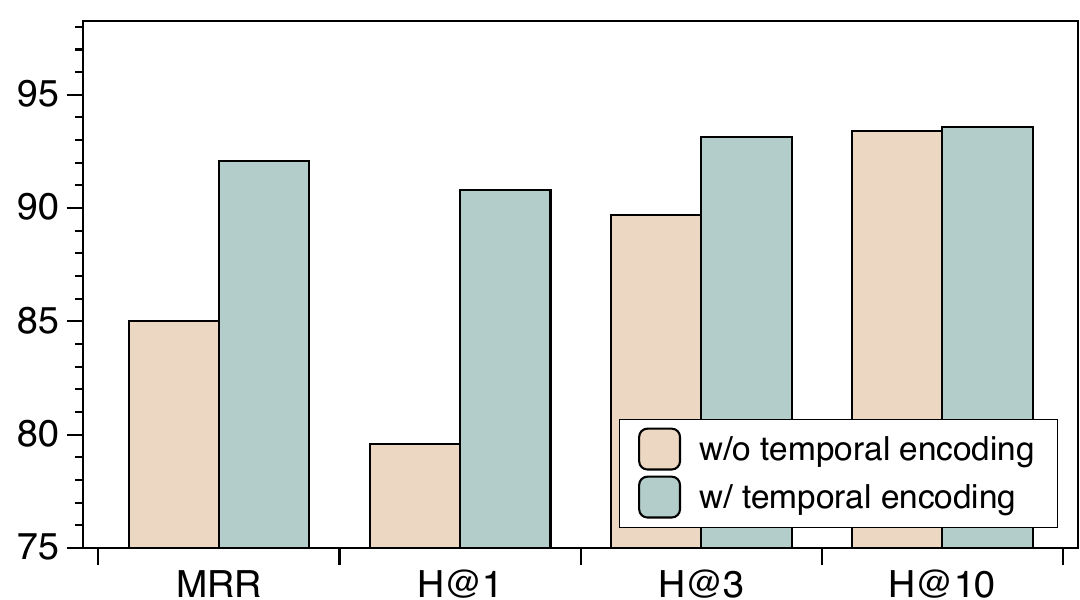}
}
\caption{Ablation Results on Temporal Encoding.}
\label{fig:ablation_time_encode}
\end{figure}

The temporal encoder $\bm\Upsilon$ plays a crucial role in modeling the temporal order of historical facts and the time factor of temporal edges in the temporal paths. Therefore, in this study, we aim to evaluate the ability of the temporal encoder in predicting future facts. To do so, we conducted an ablation analysis by removing the temporal encoder component from TiPNN and comparing its performance. The results are illustrated in Figure \ref{fig:ablation_time_encode}, where we present a performance comparison between TiPNN with and without the temporal encoder.

The experimental results demonstrate that when the temporal encoder is removed, the performance of TiPNN decreases across all datasets. This outcome aligns with our expectations since relying solely on the query-aware static representation from the temporal relation modeling cannot capture the temporal information of historical facts. As a consequence, the absence of the temporal encoder hinders the accurate learning of temporal features among past facts.

% \begin{figure}[!htbp]
% \centering
% \subfigure[ICEWS18]{
% \includegraphics[width=6.5cm]{imgs/itr/ICEWS18.pdf}
% }
% % \hspace{5mm}
% % \vspace{-3mm}
% \subfigure[YAGO]{
% \includegraphics[width=6.5cm]{imgs/itr/YAGO.pdf}
% }
% \caption{Independence Temporal Relation Encoding.}
% \label{fig:itr}
% \end{figure}

Furthermore, we also investigated the impact of the independence of parameters in the temporal encoder on all datasets. Specifically, we considered the parameters $\bm{w}*$ and $\bm{\phi}*$ in $\bm\Upsilon$ of Equation \ref{eq:time_encode}, which are responsible for modeling the temporal features of the temporal edge $(z,p_\tau,o)$, to be independently learned for each static type $p$. 
In other words, we assumed that the same relative time distance may have different temporal representations for different relation types. 
Thus, for each distinct edge type $p$, we utilized a relation-aware temporal encoder to represent the temporal features for relative time distance $\Delta\tau$. 
We conducted experiments on all datasets, as shown in Table \ref{tab:ind_tempora_encode}. We use `Shared Param.' to indicate that the parameters in $\bm\Upsilon$ are shared and not learned independently for each edge type $p$, and `Specific Param.' to indicate that the parameters are not shared and are learned independently based on the edge types.

Based on Table \ref{tab:ind_tempora_encode_1}, we observe that using shared parameters yields better performance on the ICEWS18 and GDELT datasets. Conversely, from Table \ref{tab:ind_tempora_encode_2}, we find that independently learning the parameters of the temporal encoder is more beneficial on the WIKI and YAGO datasets. We analyze this phenomenon and suggest that 
WIKI and YAGO have fewer edge types, and independently learning specific parameters for each edge type allows for easier convergence, making this approach more advantageous compared to learning shared parameters.

\begin{table}[!htbp]
\caption{Impact of the Independence Setting of Parameters in Temporal Encoder.}
\label{tab:ind_tempora_encode}
\centering
\setlength\tabcolsep{8pt}
\subtable[ICEWS18 and GDELT]{
\begin{tabular}{c|cccc|cccc}

    \toprule
    \multirow{2}{*}{Settings} &  \multicolumn{4}{c|}{ICEWS18} & \multicolumn{4}{c}{GDELT}  \\
    
    &  MRR & H@1 & H@3 & H@10 & MRR & H@1 & H@3 & H@10     \\ 
    
    \midrule
    Shared Param. & \textbf{32.17} & \textbf{22.74} & \textbf{36.24} & \textbf{50.72} & \textbf{21.17} & \textbf{14.03} & \textbf{22.98} & \textbf{34.76} \\
    Specific Param. & 30.36 & 20.80 & 34.38 & 49.29 & 21.00 & 13.88 & 22.80 & 34.60 \\
    \bottomrule
    
    \end{tabular}\label{tab:ind_tempora_encode_1}
}
\subtable[WIKI and YAGO]{
\begin{tabular}{c|cccc|cccc}

    \toprule
    \multirow{2}{*}{Settings} &  \multicolumn{4}{c|}{WIKI} & \multicolumn{4}{c}{YAGO}  \\
    
    &  MRR & H@1 & H@3 & H@10 & MRR & H@1 & H@3 & H@10     \\ 
    
    \midrule
    Shared Param. & 82.97 & 78.89 & 86.39 & 88.40 & 91.32 & 89.65 & 92.83 & 93.24 \\
    Specific Param. & \textbf{83.04} & \textbf{79.04} & \textbf{86.45} & \textbf{88.54} & \textbf{92.06} & \textbf{90.79} & \textbf{93.15} & \textbf{93.58} \\
    \bottomrule
    
    \end{tabular}\label{tab:ind_tempora_encode_2}
}
\end{table}

\subsubsection{Variants of Temporal Edge Merging Method}

The temporal edge merging operation $\textsc{Tmsg}(\cdot)$ utilizes the message passing mechanism to iteratively extend the path length while merging the query-relevant path information. As mentioned in Equation \ref{eq:tmsg}, we employ the scaling operator from DistMult~\cite{yang2014embedding} for computing the current temporal path representation with the features of temporal edges through element-wise multiplication. Additionally, we have also conducted experiments using the translation operator from TransE~\cite{bordes2013translating} (i.e., element-wise summation) and the rotation operator from RotatE~\cite{2019RotatE} to provide further evidence supporting the effectiveness of the temporal edge merging operation.

\begin{figure}[!htbp]
\centering
\subfigure[ICEWS18]{
\includegraphics[width=6.5cm]{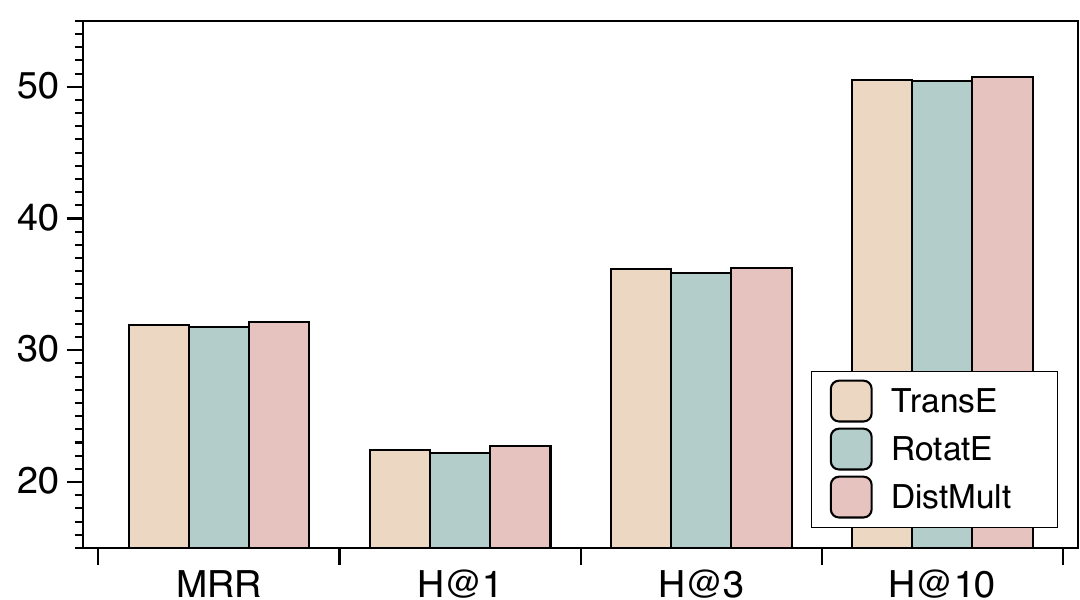}
}
% \hspace{5mm}
% \vspace{-3mm}
\subfigure[YAGO]{
\includegraphics[width=6.5cm]{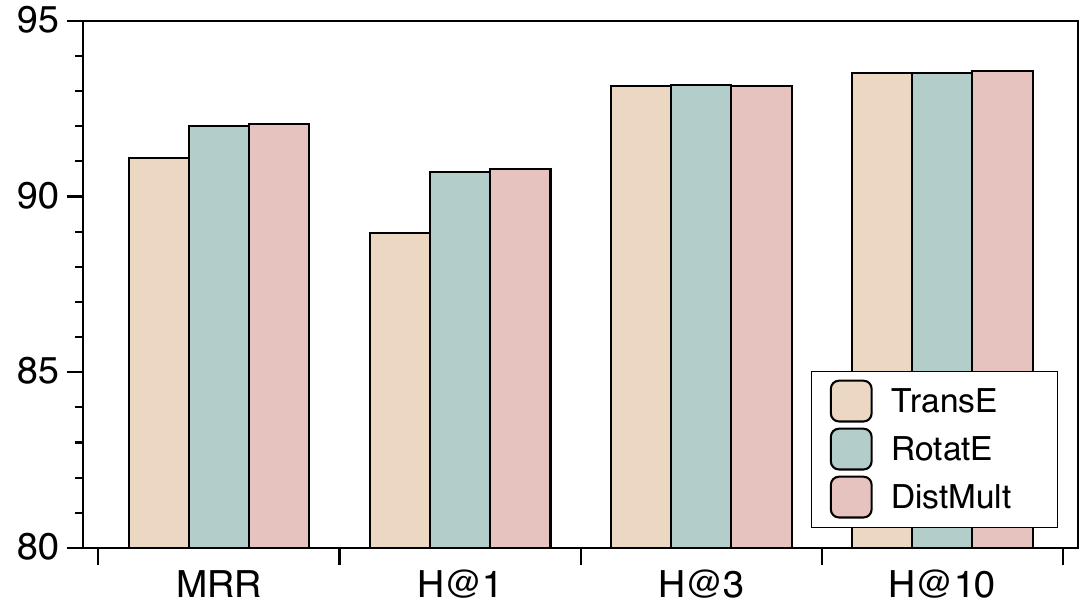}
}
\caption{Ablation Results on Temporal Edge Merging Method.}
\label{fig:ablation_tmsg}
\end{figure}

We conducted experimental validation and comparison on two types of knowledge graphs: the event-based graph ICEWS18 and the knowledge-based graph YAGO. Figure \ref{fig:ablation_tmsg} displays the results obtained with different merging operators. The findings demonstrate that TiPNN benefits from these excellent embedding methods, performing on par with DistMult and RotatE, and outperforming TransE, particularly evident on MRR and H@1 in the knowledge-based graph YAGO.

\begin{figure}[!htbp]
\centering
\subfigure[ICEWS18]{
\includegraphics[width=6.5cm]{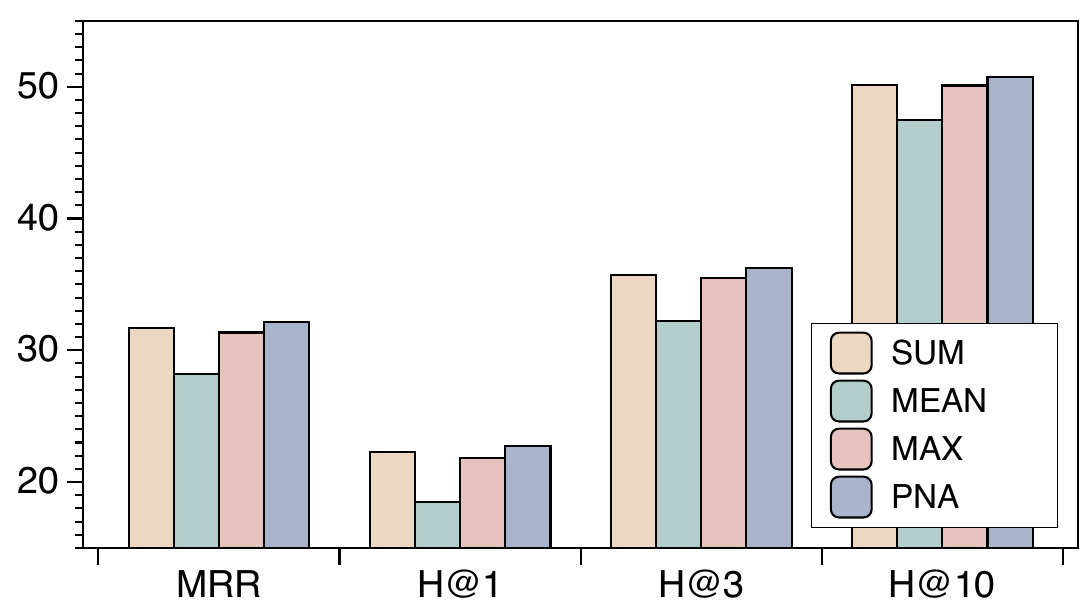}
}
% \hspace{5mm}
% \vspace{-3mm}
\subfigure[YAGO]{
\includegraphics[width=6.5cm]{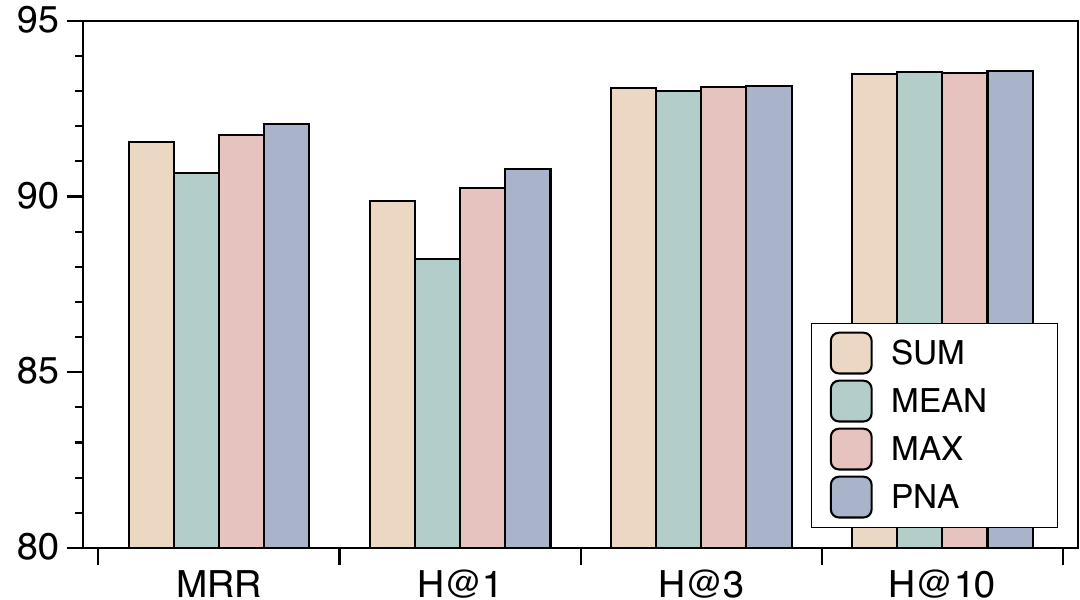}
}
\caption{Ablation Results on Path Aggregation Method.}
\label{fig:ablation_agg}
\end{figure}

\subsubsection{Variants of Path Aggregation Method}

The path aggregation operation $\textsc{Agg}(\cdot)$ is performed after merging the temporal edges at each step, and the aggregated feature serves as the temporal path representation from the query subject entity to the object entity candidates for future fact reasoning. 
As mentioned in Section \ref{sec:operating}, we employ the principal neighborhood aggregation (PNA)~\cite{corso2020principal} as the aggregator to aggregate the propagated messages. PNA aggregator considers mean, maximum, minimum, and standard deviation as aggregation features to obtain comprehensive neighbor information. To validate its effectiveness, we compare it with several individual aggregation operations: SUM, MEAN, and MAX. 

Similarly, we conduct experiments on both the event-based graph ICEWS18 and the knowledge-based graph YAGO. Figure~\ref{fig:ablation_agg} illustrates the results with different aggregation methods. The findings demonstrate that TiPNN performs optimally with the PNA aggregator and performs the worst with the MEAN aggregator, consistently in both YAGO and ICEWS18, except for the H@3 and H@10 metrics on YAGO, where the differences are less pronounced.

\subsection{Parameter Study (RQ3)}

To provide more insights on query-aware temporal path processing, we test the performance of TiPNN with different sampling history lengths $m$, and the number of temporal path aggregation layers $\omega$.

\subsubsection{Sampling History Length}\label{sec:history_len_study}

\begin{figure}[h]
\centering
\subfigure[ICEWS18]{
\includegraphics[width=6.5cm]{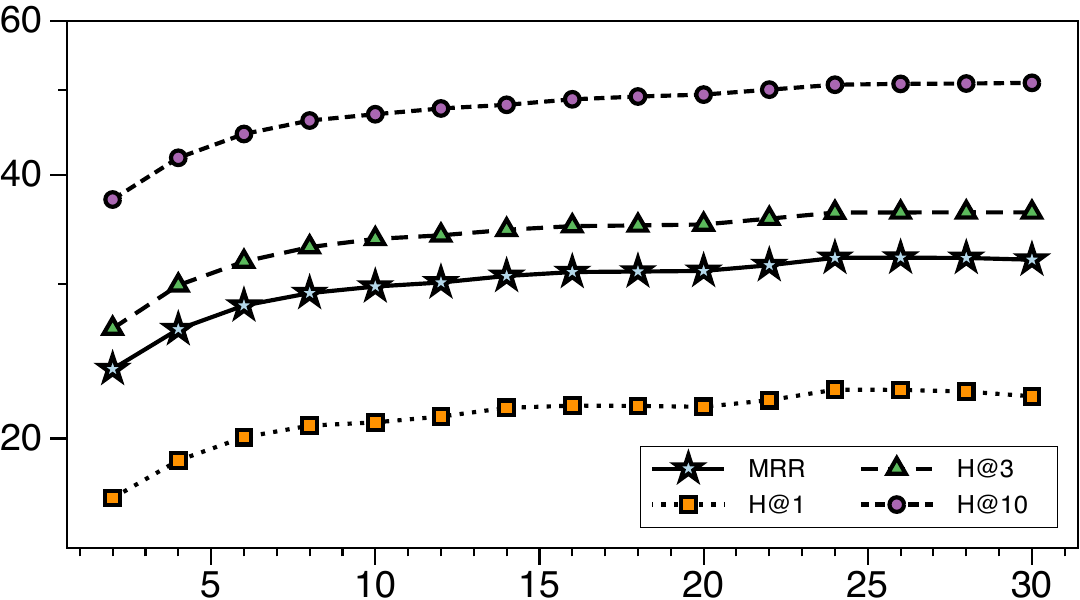}
}
% \hspace{5mm}
% \vspace{-3mm}
\subfigure[GDELT]{
\includegraphics[width=6.5cm]{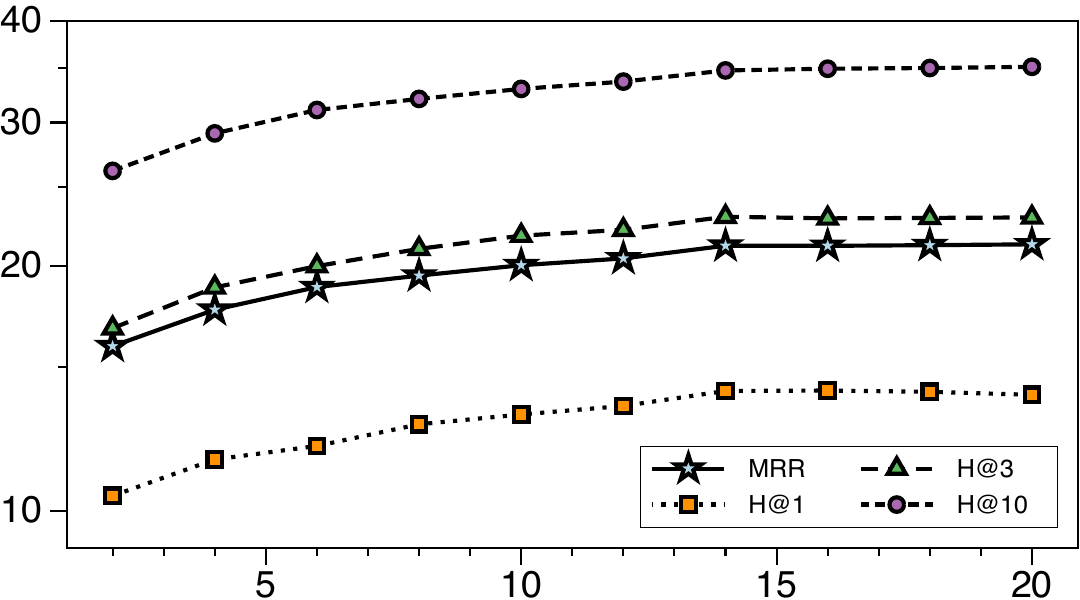}
}
\\
\subfigure[WIKI]{
\includegraphics[width=6.5cm]{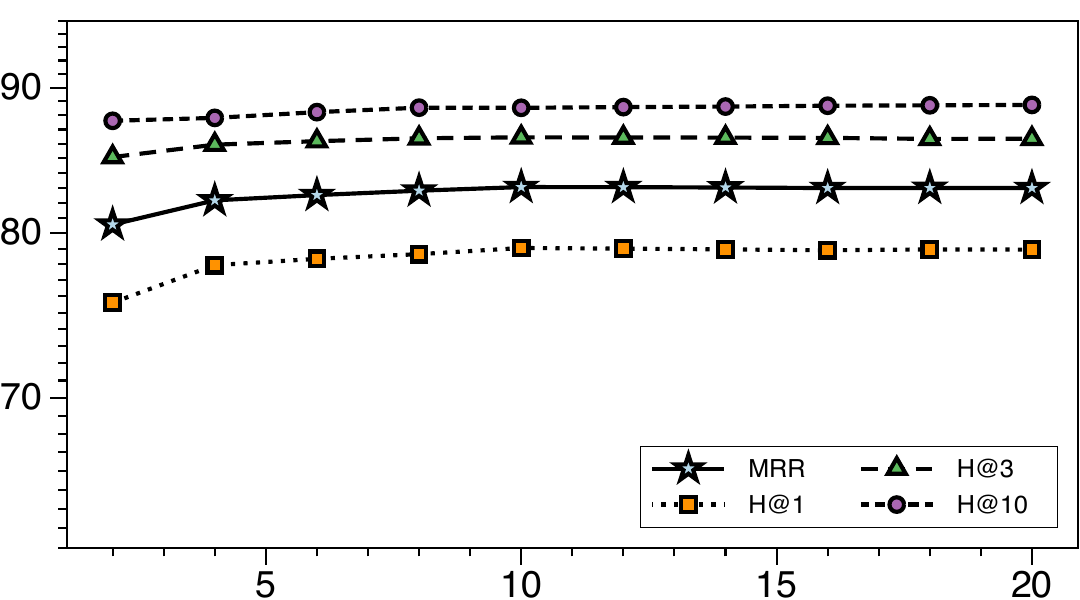}
}
% \hspace{5mm}
\subfigure[YAGO]{
\includegraphics[width=6.5cm]{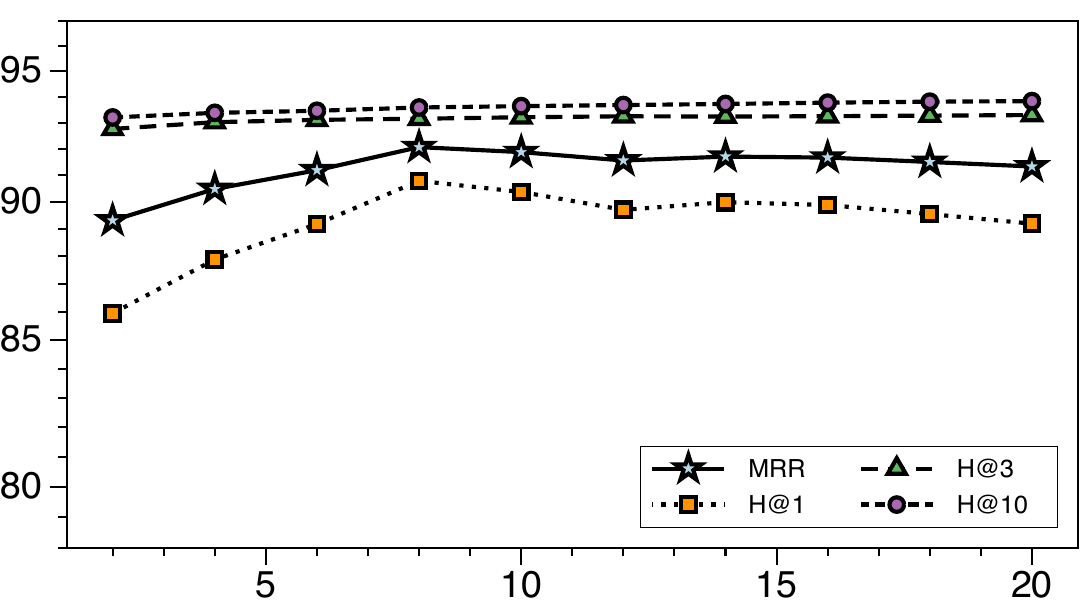}
}
\caption{The Performance of Different History Length Settings.}
\label{fig:history_length}
\end{figure}

In order to comprehensively capture the interconnection patterns between entities in the historical subgraphs, we constructed a history temporal graph to learn query-aware temporal paths between entities, enabling the prediction of future missing facts. Unlike previous methods based on subgraph sequences~\cite{li2021temporal,li-etal-2022-complex}, we fuse multiple historical subgraphs for reasoning. 
However, it may lead to the inclusion of redundant edge information from different timestamps when the historical length is excessively long, resulting in unnecessary space consumption in the learning process. Moreover, facts that are far from the prediction target timestamp have minimal contribution to the prediction, even with the addition of relative distance-aware temporal encoding in the temporal relation encoder. Conversely, if the history length is too short, it can hinder the model's ability to capture cross-time features between entities, thereby reducing the modeling performance for temporal patterns in history temporal graph. Hence, we conducted a discussion on the history length to find a balance.  The experimental results of the history length study are shown in Figure \ref{fig:history_length}.

Since TiPNN initializes the learning of temporal paths based on the query, it requires the history temporal graph's topological structure to include as many entities as possible to ensure that the learned temporal paths from the query subject entity to each candidate object entity capture more temporal connection patterns. We present the density and interval for each dataset, as shown in Table \ref{tab:density}, where $|\mathcal{V}|$ denotes the number of entities in the dataset.

\begin{table}[!htbp]
    \centering
    % \footnotesize
    \setlength\tabcolsep{18pt}
    \caption{Density and Interval of Datasets ($|\mathcal{V}_{avg}|$ is the average number of entities involved in the subgraph of each timestamp.).}
    \begin{tabular}{c|cccc}
    \toprule
    Datasets & ICEWS18 & GDELT & WIKI & YAGO  \\ 
    \midrule
    Interval & 24 hours & 15 mins & 1 year & 1 year \\  
    $|\mathcal{V}|$  & 23,033 & 7,691 & 12,554 & 10,623 \\ 
  
    % $|\mathcal{T}|$  & 304 & 2,976 & 232 & 189 \\ 
    $|\mathcal{V}_{avg}|$  & 986.44 & 393.18 & 2,817.47 & 1,190.72 \\ 
    \midrule
    $\frac{|\mathcal{V}|}{|\mathcal{V}_{avg}|}$ & 23.35 & 19.56 & 4.46 & 8.92 \\ 
    \bottomrule
    \end{tabular}

    \label{tab:density}
\end{table}

For WIKI and YAGO, which have longer time intervals (i.e., 1 year), each subgraph contains more factual information, and the average number of entities in each subgraph is naturally higher. As a result, these datasets demonstrate relatively stable performance across various history length settings, as shown in Figure \ref{fig:history_length}(c\&d). 
On the other hand, ICEWS18 and GDELT datasets exhibit different behavior, as depicted in Figure \ref{fig:history_length}(a\&b). Their time intervals are relatively shorter compared to WIKI and YAGO. Consequently, they require longer history lengths to compensate for data sparsity and incorporate more comprehensive historical information in the history temporal graph.

To assess the impact of different history lengths on the model's performance, we calculate the ratio of $|\mathcal{V}|$ to $|\mathcal{V}_{avg}|$ and present it in the last row of Table \ref{tab:density}. We observe that the model's performance tends to stabilize around this ratio, providing a useful reference for parameter tuning. Finally, we choose the balanced parameters for ICEWS18, GDELT, WIKI, and YAGO as 25, 15, 10, and 8, respectively.

\subsubsection{Number of Temporal Path Aggregation Layers}

The temporal path aggregation layer is a fundamental computational unit in TiPNN. It operates on the constructed history temporal graph using the temporal edge merging and path aggregation operation to learn query-aware temporal path representations for future fact inference. The number of layers in the temporal path aggregation layer directly impacts the number of hops in the temporal message passing on the history temporal graph, which corresponds to the logical maximum distance of temporal paths. Therefore, setting the number of layers faces a similar challenge as determining the history length in Section ~\ref{sec:history_len_study}.

When the number of layers is too high, TiPNN captures excessively long path information in temporal paths, leading to increased space consumption. However, these additional paths do not significantly improve TiPNN's performance because distant nodes connected to the target node via multiple hops are less relevant for inference of future facts. Therefore, considering distant nodes in the inference process has limited impact on the accuracy of predictions for future facts. On the other hand, if the number of layers is too low, TiPNN may not effectively capture long-distance logical connections in the history temporal graph. This may result in insufficient learning of comprehensive topological connection information and precise combination rules of temporal paths.

\begin{figure}[!htbp]
\centering
\subfigure[ICEWS18]{
\includegraphics[width=6.5cm]{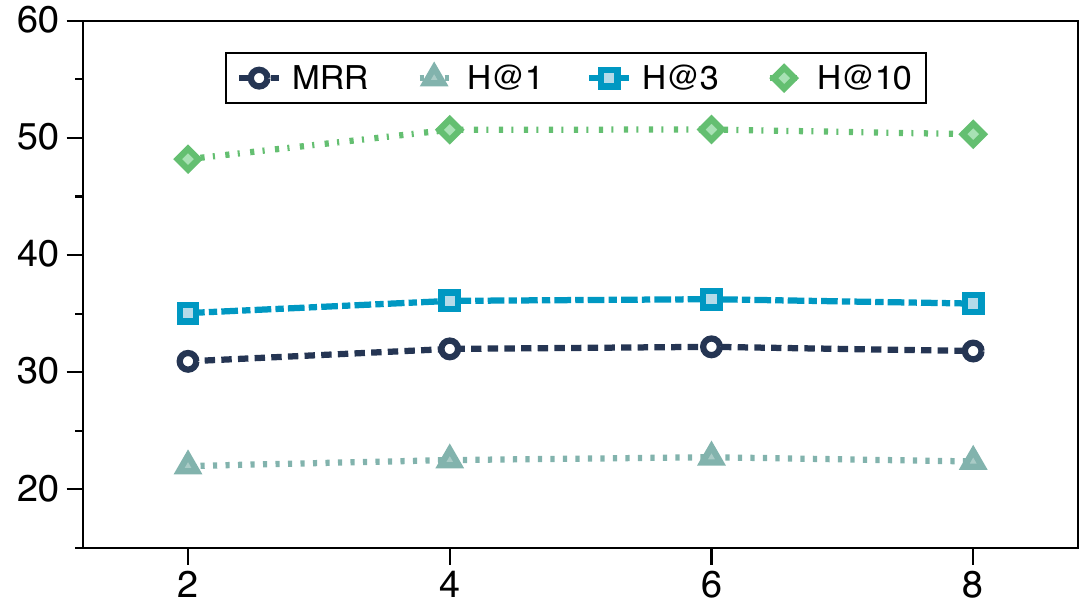}
}
% \hspace{5mm}
% \vspace{-3mm}
\subfigure[YAGO]{
\includegraphics[width=6.5cm]{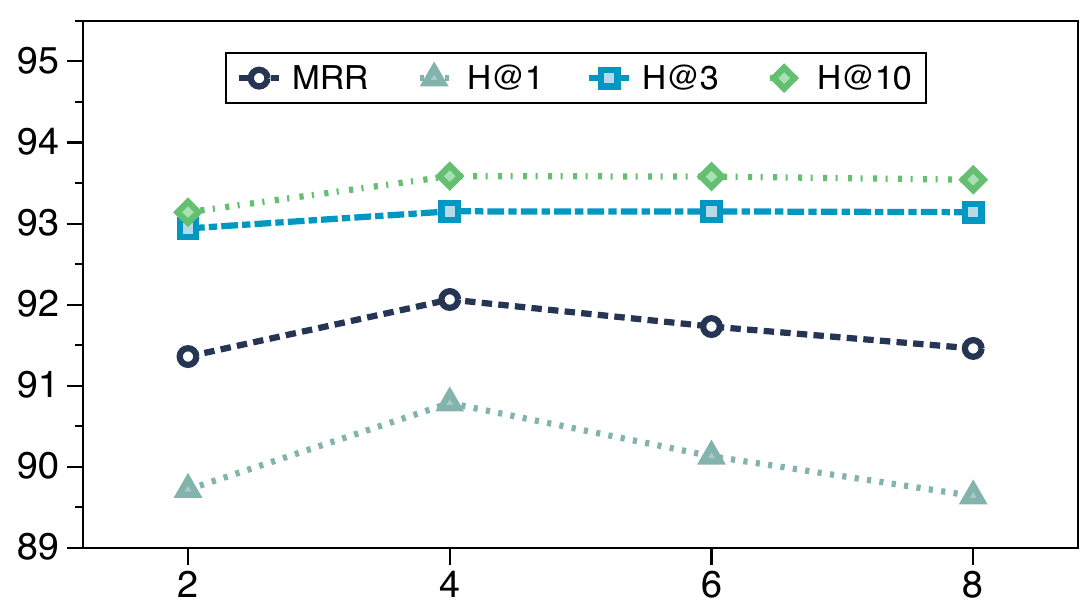}
}
\caption{The Performance of Different Numbers of Aggregation Layers.}
\label{fig:num_agg_layer}
\end{figure}

To find an optimal setting for the number of layers in temporal path aggregation, we conducted experiments on ICEWS18 and YAGO as shown in Figure \ref{fig:num_agg_layer}, which illustrates the results for varying the number of temporal path aggregation layers. 
The results indicate that ICEWS18 performs best with 6 layers, and the number of layers has a minor impact on the results. As the number of layers increases, there is no significant improvement in the model's accuracy. On the other hand, YAGO performs best with 4 layers, and beyond 4 layers, the model's performance declines noticeably. We consider that this difference in performance could be attributed to the fact that ICEWS18, as an event-based graph, involves longer-distance logical connections for inferring future facts. On the contrary, YAGO, being a knowledge-based graph, has sparser relation types, resulting in more concise reasoning paths during inference. We finally set the optimal number of layers for ICEWS18 and GDELT to 6, and 4 for WIKI and YAGO.

\subsection{Comparison on Prediction Time (RQ4)}

\begin{figure*}[!htbp]
    \centering
    \includegraphics[width=0.8\textwidth]{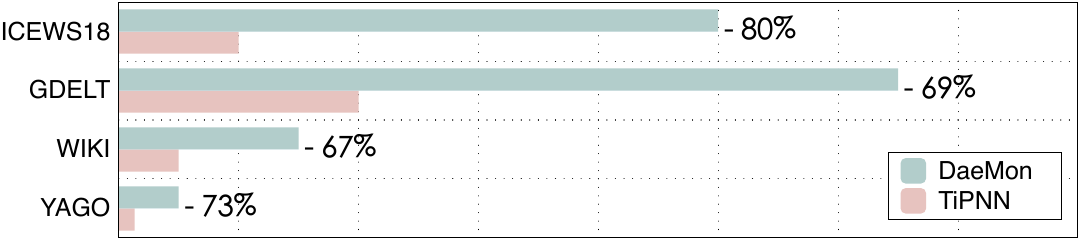}
    \caption{Runtime Comparison (The runtime is proportionally represented as multiples of the unit time.).}
    \label{fig:prediction_time}

\end{figure*}

To further analyze the efficiency of TiPNN, we compared the inference time of TiPNN with DaeMon \cite{dong2023adaptive} on the TKG reasoning task. For a fair comparison, we use the test sets of four datasets and align the model parameters under the same setting and environment. The runtime comparison between the two models is illustrated in Figure \ref{fig:prediction_time}, where we present the ratio of their inference times in terms of multiples of the unit time.
The comparative experiments demonstrate that TiPNN achieves significant reductions in runtime compared to DaeMon, with time savings of approximately 80\%, 69\%, 67\%, and 73\% on the ICEWS18, GDELT, WIKI, and YAGO datasets, respectively. 

The efficiency of TiPNN is mainly attributed to its construction of history temporal graphs, which enables the modeling of logical features between temporal facts and the capture of semantic and temporal patterns from historical moments. In contrast, DaeMon relies on graph evolution methods to handle the potential path representations, while TiPNN employs a reduced number of message-passing layers to learn real existing path features in the history temporal graph. Additionally, TiPNN leverages temporal edges to simultaneously capture relational and temporal features, eliminating the need for separate sequence modeling units to learn temporal representations. As a result, TiPNN exhibits lower complexity and higher learning efficiency compared to DaeMon.

\subsection{Inductive Setting (RQ5)}
To validate the model's inductive reasoning ability, we also design experiments for TiPNN in an inductive setting.
The inductive setting is a common scenario in knowledge graphs, where during the training process, the model can only access a subset of entities, and during testing, it needs to reason about unseen entities~\cite{Teru2020InductiveRP}. This setting aims to simulate real-world situations where predictions are required for new facts based on existing knowledge.
In the context of the temporal knowledge graph, the inductive setting requires the model to perform reasoning across time, meaning that the learned representations from the historical subgraphs should generalize to future timestamps for predicting missing facts. This demands the model to possess the ability to generalize to unseen entities, leveraging the learned temporal connectivity patterns from the historical subgraphs.

In the inductive setting experiment, we train the model using only a portion of entities and then test its inference performance on the reasoning of the unseen entities. This type of validation allows for a comprehensive evaluation of the model's generalization and inductive reasoning capabilities, validating the effectiveness of TiPNN in predicting future missing facts. 

In the context of TKGs, there are few existing datasets suitable for inductive validation. Therefore, we follow the rules commonly used in KG for inductive settings and construct an inductive dataset specifically tailored for TKGs~\cite{Teru2020InductiveRP}. Finally, we conduct experiments on our proposed model using the inductive dataset to evaluate its inference performance under the inductive setting.

\subsubsection{Inductive Datasets}

In order to facilitate the inductive setting, we create a fully-inductive benchmark dataset by sampling disjoint entities from the YAGO dataset~\cite{2013YAGO3}. Specifically, the inductive dataset consists of a pair of TKGs: YAGO$^1$ and YAGO$^2$, satisfying the following conditions: (i) they have non-overlapping sets of entities and (ii) they share the same set of relations. 
To provide a comprehensive evaluation, we have sampled three different versions of the pair inductive dataset based on varying proportions of entity set cardinality. Table \ref{tab:ind_data_stat} provides the statistical data for these inductive datasets, where $|\mathcal{V}|$ denotes the number of entities in the dataset, $|\mathcal{R}|$ denotes the number of relations in the dataset. The labels v1(5:5), v2(6:4), and v3(7:3) in the table correspond to the three different datasets created with different entity set partition ratios.

\begin{table}[!htbp]
    \centering
    \setlength\tabcolsep{10pt}
    \caption{Statistics of Inductive Datasets ($\mathcal{E}_{train}$, $\mathcal{E}_{valid}$, $\mathcal{E}_{test}$ are the numbers of facts in training, validation, and test sets.).}
    \begin{tabular}{cccccccc}
    \toprule
    \multicolumn{2}{c}{Datasets} & \multicolumn{1}{c}{$|\mathcal{V}|$} & \multicolumn{1}{c}{$|\mathcal{R}|$} & \multicolumn{1}{c}{$\mathcal{E}_{train}$} & \multicolumn{1}{c}{$\mathcal{E}_{valid}$} & \multicolumn{1}{c}{$\mathcal{E}_{test}$} & Interval \\ 
    \midrule
    \multirow{2}{*}{v1 (5:5)}  & YAGO$^1$ & 3,980                   & 10                      & 39,588                     & 4,681                      & 6,000                     & 1year    \\
                         & YAGO$^2$ & 3,963                   & 10                      & 37,847                     & 4,544                      & 5,530                     & 1year    \\ 
    \midrule
    \multirow{2}{*}{v2 (6:4)}  & YAGO$^1$ & 4,590                   & 10                      & 46,070                     & 5,685                      & 5,904                     & 1year    \\
                         & YAGO$^2$ & 3,293                   & 10                      & 33,091                     & 4,836                      & 4,244                     & 1year    \\ 
    \midrule
    \multirow{2}{*}{v3 (7:3)}  & YAGO$^1$ & 5,705                   & 10                      & 64,036                     & 7,866                      & 9,663                     & 1year    \\
                         & YAGO$^2$ & 2,407                   & 10                      & 20,884                     & 2,545                      & 3,180                     & 1year    \\ 
    \bottomrule
    \end{tabular}
    \label{tab:ind_data_stat}
\end{table}

\subsubsection{Results of Inductive Experiment}

In the Inductive setting, we conduct inference experiments and validations by cross-utilizing the training and test sets of a pair of TKGs: YAGO$^1$ and YAGO$^2$. 
Specifically, we train on the training set of YAGO$^1$ and test on the test set of YAGO$^2$, and vice versa, to achieve cross-validation in each version of the dataset. 
Additionally, we include experimental results in the transductive setting as a baseline for comparison with the inductive setting. 
We used `bias' to display the difference in results between the two settings. As shown in Table \ref{tab:inductive_result}, the experimental results demonstrate that the bias is within a very small range.

The success of TiPNN in handling the Inductive setting is not surprising. This is because TiPNN models the structural semantics and temporal features of historical facts without relying on any learnable entity-related parameters. Instead, it leverages temporal edge features to model temporal path features relevant to the query. This inherent capability of TiPNN allows it to naturally perform inference on datasets that include unseen entities, making it well-suited for the inductive setting.

\begin{table}[!htbp]
    \centering
    \setlength\tabcolsep{3pt}
    \caption{Inductive Performance of TiPNN ($\rightarrow$ points from training set to test set corresponding to the inductive setting, and the row results without arrow marked is the corresponding transductive setting as the baseline. `Bias' denotes represents the difference in comparison.).}
    \begin{tabular}{c|cccc|cccc|cccc}

    \toprule
    \multirow{2}{*}{Dataset} &  \multicolumn{4}{c|}{v1} & \multicolumn{4}{c|}{v2} & \multicolumn{4}{c}{v3}  \\
    
    &  MRR & H@1 & H@3 & H@10 & MRR & H@1 & H@3 & H@10 & MRR & H@1 & H@3 & H@10    \\ 
    
    \midrule
    YAGO$^1$                     & 90.85 & 89.31 & 92.10 & 92.58 & 91.19 & 89.79 & 92.35 & 92.73 & 91.67 & 90.46 & 92.72 & 93.09 \\ 
    YAGO$^2\rightarrow$YAGO$^1$  & 90.75 & 89.09 & 92.11 & 92.64 & 91.07 & 89.60 & 92.27 & 92.73 & 91.19 & 89.58 & 92.62 & 92.94 \\
    % \midrule
    \multirow{1}{*}{Bias ($\pm$)} &  \multicolumn{4}{c|}{$\leq$ 0.22} & \multicolumn{4}{c|}{$\leq$ 0.19} & \multicolumn{4}{c}{$\leq$ 0.88}  \\ 

    \midrule
    YAGO$^2$                    & 91.33 & 89.83 & 92.61 & 93.02 & 92.88 & 91.61 & 93.93 & 94.27 & 90.66 & 89.22 & 91.70 & 92.43 \\ 
    YAGO$^1\rightarrow$YAGO$^2$ & 91.15 & 89.47 & 92.65 & 93.07 & 92.88 & 91.63 & 93.90 & 94.36 & 90.96 & 89.66 & 91.95 & 92.63 \\ 
    % \midrule
    \multirow{1}{*}{Bias ($\pm$)} &  \multicolumn{4}{c|}{$\leq$ 0.36} & \multicolumn{4}{c|}{$\leq$ 0.09} & \multicolumn{4}{c}{$\leq$ 0.44}  \\ 

    \bottomrule
    \end{tabular}
    \label{tab:inductive_result}

\end{table}

\subsection{Reasoning Evidence (RQ6)}

Since we have integrated historical information into the constructed history temporal graph and perform inference on future facts by modeling temporal paths, we can use the path in the history temporal graph to provide evidence for the inference process and visualize the reasoning basis.
% making it easier for users to understand the results and fine-tune the model if needed. 
An intuitive idea is that the model should provide important reasoning paths from the history temporal graph that contribute significantly to the inference for the corresponding response. 
Although the temporal path representation we obtain is a comprehensive representation logically aggregated from multiple paths in the history temporal graph, we can still estimate the importance of each individual path following the local interpretation method~\cite{cf736d955d6e4296a9b7255bfee3b403,10.1007/978-3-319-10590-1_53}. 
As described in Section~\ref{sec:score_func}, we convert the learned temporal path representation into corresponding scores, and thus, we can estimate the importance of paths through backtracking.

\begin{table}[!h]
    \centering
    \small
    \caption{Reasoning Evidence of Responses to Selected Queries from ICEWS18. The queries, responses, and the corresponding top-2 reasoning evidence with importance (higher values indicate higher importance) are provided. The attached superscript $^{-1}$ denotes the inverse relations. The timestamp in the quadruple is converted into mm/dd/yyyy format for representation.}
    \begin{tabular}{rl}
    
    \toprule
    \textbf{Query:}      &   \textbf{(Shinzo Abe, Make a visit, ?, 10/23/2018)} \\
    \textbf{Response:}   &   China   \\
    \midrule
    0.784  &   $<$Shinzo Abe, Express intent to meet or negotiate, China, 10/22/2018$>$. \\
    \rule{0pt}{1mm}   \\
    0.469  &   $<$Shinzo Abe, Express intent to meet or negotiate, Li Keqiang, 10/21/2018$>$$\rightarrow$ \\
	~  &   $<$Li Keqiang, Express intent to meet or negotiate$^{-1}$, Shinzo Abe, 10/10/2018$>$$\rightarrow$ \\
	~  &   $<$Shinzo Abe, Make a visit, China, 10/22/2018$>$. \\

    \midrule
    \textbf{Query:}      &   \textbf{(European Union, Engage in diplomatic cooperation, ?, 10/23/2018)} \\
    \textbf{Response:}   &   United Kingdom   \\
    \midrule
    1.613  &   $<$European Union, Engage in diplomatic cooperation, United Kingdom, 10/14/2018$>$. \\
    \rule{0pt}{1mm}   \\
    1.044  &   $<$European Union, Express intent to engage in diplomatic cooperation (such as policy support),\\
        ~  &   United Kingdom, 10/22/2018$>$. \\

    \midrule
    \textbf{Query:}      &   \textbf{(Russia, Host a visit, ?, 10/24/2018)} \\
    \textbf{Response:}   &   Head of Government (Italy)   \\
    \midrule
    2.350  &   $<$Russia, Make a visit$^{-1}$, Head of Government (Italy), 10/23/2018$>$. \\
    \rule{0pt}{1mm}   \\
    1.338  &   $<$Russia, Express intent to meet or negotiate$^{-1}$, Head of Government (Italy), 10/22/2018$>$. \\

    \midrule
    \textbf{Query:}      &   \textbf{(Citizen (Thailand), Threaten with military force, ?, 10/25/2018)} \\
    \textbf{Response:}   &   Police (Thailand)   \\
    \midrule
    0.951  &   $<$Citizen (Thailand), fight with small arms and light weapons$^{-1}$, Police (Thailand), 10/13/2018$>$. \\
    \rule{0pt}{1mm}   \\
    0.376  &   $<$Citizen (Thailand), Accuse, Thailand, 10/03/2018$>$$\rightarrow$ \\
        ~  &   $<$Thailand, Return, release person(s), Citizen (Thailand), 10/10/2018$>$$\rightarrow$ \\
        ~  &   $<$Citizen (Thailand), fight with small arms and light weapons$^{-1}$, Police (Thailand), 10/13/2018$>$. \\

    \midrule
    \textbf{Query:}      &   \textbf{(Health Ministry (India), Criticize or denounce$^{\bm{-1}}$, ?, 10/27/2018)} \\
    \textbf{Response:}   &   Citizen (India)   \\
    \midrule
    0.757  &   $<$Health Ministry (India), Make an appeal or request, Citizen (India), 10/26/2018$>$. \\
    \rule{0pt}{1mm}   \\
    0.367  &   $<$Health Ministry (India), Make statement, Government (India), 10/19/2018$>$$\rightarrow$ \\
	~  &   $<$Government (India), Criticize or denounce$^{-1}$, Citizen (India), 10/26/2018$>$. \\

    \bottomrule
    \end{tabular}
    \label{tab:evidence}
\end{table}

Drawing inspiration from path interpretation~\cite{zhu2022neural}, we define the importance of a path as the weights assigned to it during the iteration process. 
We achieve this by computing the partial derivative of the temporal path scores with respect to the path's weights. Specifically, for a reasoning response $(s,r,o,t+1)$ of a missing object query $(s,r,?,t+1)$, we consider the top-k paths as the basis for inference, which is defined as shown in Equation \ref{eq:evidence}. 

\begin{equation}\label{eq:evidence}
    \text{P}_1, \text{P}_2, \cdots, \text{P}_k = 
    \underset{\text{P} \in \mathcal{P}_{s \rightarrow o}^{\prec t+1}}{\text{Top-}k}
    \frac{\partial\ p(s,r,o)}{\partial\ P}  
\end{equation}

In practice, since directly computing the importance of an entire path is facing a challenge, we calculate the importance of each individual temporal edge in the history temporal graph, which can be obtained using automatic differentiation. And then, we aggregate the importance of temporal edges with a summation operation within each path to determine the corresponding path's importance. By applying beam search to traverse and compute the importance of each path, we can finally obtain the top-k most important paths as our reasoning evidence.

We selected several queries from the test set of event-based graph ICEWS18 to conduct an analysis and discussion of the reasoning evidence. As shown in Table \ref{tab:evidence}, it presents the responses corresponding to these queries and their top-2 related reasoning evidence. 
Here, we provide our analysis and interpretation of the reasoning evidence for the first three queries.
(I) For the first query \textit{(Shinzo Abe, Make a visit, ?, 10/23/2018)}, TiPNN's response is China. The most significant reasoning clue is that Shinzo Abe expressed intent to meet or negotiate on 10/22/2018. Additionally, on 10/21/2018, there was also an expression of intent to meet or negotiate, followed by the action of \textit{Make a visit} on 10/22/2018. These two clues are logically consistent and form a reasoning pathway: \textit{Express intent to meet or negotiate $\rightarrow$ Make a visit}.
(II) For the query \textit{(European Union, Engage in diplomatic cooperation, ?, 10/23/2018)}, the response is United Kingdom. One of the reasoning clues is that on 10/14/2018, European Union engaged in diplomatic cooperation with the United Kingdom, and on 10/22/2018 (the day before the query time), European Union expressed intent to engage in diplomatic cooperation with the United Kingdom. This is easily understandable as it follows a logical reasoning pathway: \textit{Express intent to engage in diplomatic cooperation $\rightarrow$ Engage in diplomatic cooperation}.
(III) For the query \textit{(Russia, Host a visit, ?, 10/24/2018)}, the response is Head of Government (Italy). One reliable reasoning clue is that on 10/23/2018, Russia was visited by the Head of Government (Italy). This is also not surprising, as it follows a logical reasoning pathway: \textit{Make a visit$^{-1}$ $\rightarrow$ Host a visit}, which is in line with common sense rules.

Through these examples of reasoning evidence, we can observe that TiPNN is capable of learning temporal reasoning logic from the constructed history temporal graphs. These reasoning clues provide users with more comprehensible inference results and intuitive evaluation criteria. One can utilize these clues to understand the reasoning process for future temporal facts, thereby increasing confidence in the inference results and enabling potential refinements or improvements when needed.

\section{Related Work}
We divide the related work into two categories: (i) knowledge graph reasoning methods, and (ii) temporal knowledge graph reasoning methods.
\subsection{Knowledge Graph Reasoning}
In recent years, knowledge graph representation learning has witnessed significant developments, aiming to embed entities and relations into continuous vector spaces while capturing their semantics~\cite{bordes2013translating,2019RotatE,li-etal-2022-transher}. These methods can be broadly categorized into three categories: translation-based methods, semantic matching methods, and neural network-based methods.
Translation-based methods, such as TransE~\cite{bordes2013translating}. It considers relations as translations from subject entities to object entities in the vector space. Building upon TransE, several improved methods have been proposed, including TransH~\cite{wang2014knowledge}, TransR~\cite{lin2015learning}, TransD~\cite{ji2015knowledge} and TransG~\cite{xiao2015transg}, which introduce various strategies to enhance the modeling of entity-relation interactions.
For semantic matching methods, RESCAL~\cite{nickel2011three} proposes a tensor-based relational learning approach capable of collective learning. DistMult~\cite{yang2014embedding} simplifies RESCAL using diagonal matrices for efficiency, while HoIE~\cite{nickel2016holographic} and ComplEx~\cite{trouillon2016complex} extend the representation capacity by incorporating higher-order interactions and complex-valued embeddings~\cite{liu2020representation}, respectively.
Neural network-based methods have also gained attention, with approaches like GCN~\cite{welling2016semi}, R-GCN~\cite{schlichtkrull2017modeling}, WGCN~\cite{shang2019end}, VR-GCN~\cite{ye2019vectorized}, and CompGCN~\cite{vashishth2019composition}, which integrate content and structural features within the graph, allowing for joint embedding of entities and relations in a relational graph~\cite{Ning2021LightCAKEAL,qiao2020context}. These methods effectively capture complex patterns and structural dependencies in KGs, pushing the boundary of KG representation learning.
Despite the successes of existing methods in reasoning with static KGs, they fall short when it comes to predicting temporal facts due to the lack of temporal modeling.

\subsection{Temporal Knowledge Graph Reasoning}
An increasing number of studies have started focusing on the representation learning of temporal knowledge graphs, aiming to consider the temporal ordering of facts and capture the temporal knowledge of events. Typically, TKG reasoning methods can be categorized into two main classes based on the range of query timestamps: interpolation reasoning and extrapolation reasoning.

For interpolation reasoning, the objective is to infer missing facts from the past within observed data~\cite{dasgupta2018hyte,esteban2016predicting,A2018Learning}. 
TTransE~\cite{2018Deriving} extends the TransE~\cite{bordes2013translating} by incorporating relations and timestamps as translation parameters between entities. 
TA-TransE~\cite{garcia2018learning} integrates the timestamps corresponding to fact occurrences into the relation representations, 
while HyTE~\cite{dasgupta2018hyte} associates timestamps with corresponding hyperplanes. 
Additionally, based on ComplEx~\cite{trouillon2016complex}, TNTComplEx~\cite{2020Tensor} adopts a unique approach where the TKGs are treated as a 4th-order tensor, and the representations are learned through canonical decomposition. 
However, these methods are not effectively applicable for predicting future facts~\cite{2020Diachronic,han2020dyernie,2018Deriving,sadeghian2016temporal}.

In contrast, the extrapolation setting, which this work focuses on, aims to predict facts in future timestamps based on historical TKG sequences. 
Know-Evolve~\cite{trivedi2017know} uses the temporal point process to represent facts in the continuous time domain. However, it falls short of capturing long-term dependencies. 
Similarly, DyREP~\cite{trivedi2019dyrep} posits representation learning as a latent mediation process and captures the changes of the observed processes.
CyGNet~\cite{zhu2021learning} utilizes a copy-generation mechanism that collects frequently repeated events with the same subject entities and relations to the query for inferring.
RE-NET~\cite{jin2020Renet} conducts GRU and GCN to capture the temporal and structural dependencies in a sequential manner.
RE-GCN~\cite{li2021temporal} considers both structural dependencies and static properties of entities at the same time, modeling in evolving manner.
TANGO~\cite{han2021learning} designs neural ordinary differential equations to represent and model TKGs for continuous-time reasoning.
% xERTE~\cite{han2020explainable} offers an explainable model for their predictions by searching in the subgraph with attentive propagation.
% In addition, reinforcement learning methods are used to uncover the most related facts to answer the query, such as TITer~\cite{sun2021timetraveler}. 
Most existing methods are primarily transductive in nature, only focusing on modeling the graph features based on entities while overlooking the intrinsic temporal logical rules within TKG reasoning. 
As an illustration, xERTE~\cite{han2020explainable} assumes that relations do not evolve and constructs an inference graph by sampling nodes from historical subgraphs, learning time-aware entity embeddings on it, yet neglecting complex relationship patterns.
From another perspective, some path-based reasoning methods were proposed, which not only provide a new perspective for TKG reasoning but also serve as a reference for this work. For example, TITer~\cite{sun2021timetraveler} augments the historical subgraph sequence by adding auxiliary edges and employs reinforcement learning methods to guide path searching for inference.
As a result, there is significant room for development in TKG reasoning, and a need for more efficient methods that can address the robustness and scalability requirements in real-world situations.

\section{Conclusion}

In this work, we have introduced TiPNN, an innovative query-aware temporal path reasoning model for TKGs reasoning tasks, addressing the challenge of predicting future missing temporal facts under the temporal extrapolated setting. 
First, we presented a unified graph, namely the history temporal graph, which represents the comprehensive features of the historical context. This constructed graph allows for a more comprehensive utilization of temporal features between historical facts during graph representation learning.
Second, we introduced a novel concept of temporal paths, designed to capture query-relevant logical semantic paths on the history temporal graph, which can provide rich structural and temporal context for reasoning tasks.
Third, a query-aware temporal path processing framework was also designed, integrated with the introduced temporal edge merging and path aggregation functions. It enables the modeling of temporal path features over the history temporal graph for future temporal fact reasoning.

Overall, the proposed model avoids the need for separate graph learning on each temporal subgraph, making use of the unified graph to represent the information of historical feature to enhance the efficiency of the reasoning process. 
By starting from the query, capturing and learning over query-aware temporal paths, TiPNN accounts for both structural information and temporal dependencies between entities of separate subgraphs in historical context, and achieves the prediction of future missing facts while offering interpretable reasoning evidence that facilitates users' analysis of results and model fine-tuning.
Besides, in learning historical patterns, the modeling process adopts an entity-independent manner, that means TiPNN doesn't rely on specific entity representations, enabling it to naturally handle TKG reasoning tasks under the inductive setting. 
We have conducted extensive experiments on TKG reasoning benchmark datasets to evaluate the performance of TiPNN. The results demonstrate that the proposed model exhibits superior effectiveness and attains new state-of-the-art achievements.

The path-based, entity-independent approach employed in TiPNN is well-suited for TKG reasoning tasks as it explores interaction patterns among historical entities, deducing logical semantic interactions that occurred in the past. Although this approach has been applied to traditional KG reasoning ~\cite{zhu2021neural}, the explicit presence of time information and implicit temporal patterns in TKGs renders many excellent KG methods unsuitable for practical applications with a temporal dimension. Moreover, TiPNN consolidates historical subgraph sequences into a unified graph, allowing for more comprehensive utilization of interactions among entities at different times, which helps overcome the limitations and complexities associated with traditional history-evolution methods~\cite{li2021temporal,dong2023adaptive} when modeling temporal features. While TiPNN provides a novel reference for the TKG reasoning domain, several challenges remain for further research.
First, current TKG research is constrained to validation on small-scale datasets, and the performance potential for large-scale datasets is yet to be fully explored. Second, TKGs can be applied to a variety of practical scenarios such as event analysis and trajectory prediction, offering meaningful extensions beyond their current scope. Third, despite the outstanding performance and maturity of existing KG reasoning methods, efficiently integrating advanced KG techniques into the TKG domain poses a significant challenge.

\section*{Acknowledgments}
This research is funded by 
the Natural Science Foundation of China under Grant No. 61836013, 
the Science and Technology Development Fund (FDCT), Macau SAR (file no. 0014/2022/AFJ, 0123/2023/RIA2, 001/2024/SKL), 
the Start-up Research Grant of University of Macau (File no. SRG2021-00017-IOTSC),
the Postdoctoral Fellowship Program of CPSF (No.GZC20232736), 
the China Postdoctoral Science Foundation Funded Project (No.2023M743565).



\bibliographystyle{elsarticle-num} 
\bibliography{ref}





% \end{thebibliography}

\end{document}